\documentclass{article}
\usepackage{graphicx}
\usepackage{multirow}
\usepackage{mathtools}
\usepackage[preprint, nonatbib]{neurips_2020}
\usepackage{amsthm}

\usepackage[utf8]{inputenc} % allow utf-8 input
\usepackage[T1]{fontenc}    % use 8-bit T1 fonts
\usepackage{hyperref}       % hyperlinks
\usepackage{url}            % simple URL typesetting
\usepackage{booktabs}       % professional-quality tables
\usepackage{amsfonts}       % blackboard math symbols
\usepackage{nicefrac}       % compact symbols for 1/2, etc.
\usepackage{microtype}      % microtypography
\usepackage{import}
\usepackage{amsmath}
\usepackage{wrapfig}

\newcommand{\TODO}[1]{}
\renewcommand{\TODO}[1]{{\color{red} TODO: {#1}}}

\newcommand{\comment}[1]{}
\newtheorem{theorem}{Theorem}
\newtheorem{definition}[theorem]{Definition}
\newtheorem{lemma}{Lemma}
\newcommand{\R}{\mathbb{R}}
\newcommand{\Sbb}{\mathbb{S}}
\newcommand{\E}{\mathbb{E}}
\newcommand{\Tbb}{\mathbb{T}}
\newcommand{\maxl}{\max\limits_}
\newcommand{\Star}{S^{*}}

\newcommand{\ie}{\emph{i.e.}}

\newcommand{\mkd}{\mbox{MKD}}
\usepackage{xcolor}

\usepackage{subfiles}
\usepackage{subcaption}

\title{Marginal Contribution Feature Importance - an Axiomatic Approach for The Natural Case}

\author{%
  Amnon Catav\\
  School of Computer Science\\
  Tel-Aviv University\\
  \texttt{amnoncatav@mail.tau.ac.il} \\
  \And
   Boyang Fu \\
   Department of Computer Science \\
   University of California, Los Angeles \\
   \texttt{boyang1995@g.ucla.edu} \\
   \And
   Jason Ernst \\
   Department of Biological Chemistry \\
   Department of Computational Medicine \\
   Department of Computer Science \\
   University of California, Los Angeles \\
   \texttt{jason.ernst@ucla.edu}
   \And
   Sriram Sankararaman \\
   Department of Human Genetics \\
   Department of Computational Medicine \\
   Department of Computer Science \\
   University of California, Los Angeles \\
   \texttt{sriram@cs.ucla.edu} \\
  \And
  Ran Gilad-Bachrach\\
  Department of Biomedical Engineering\\
  Tel-Aviv University\\
  \texttt{rgb@tauex.tau.ac.il} \\
}

\begin{document}

\maketitle
 
\begin{abstract}
When training a predictive model over medical data, the goal is sometimes to gain insights about a certain disease. In such cases, it is common to use feature importance as a tool to highlight significant factors contributing to that disease. As there are many existing methods for computing feature importance scores, understanding their relative merits is not trivial. Further, the diversity of scenarios in which they are used lead to different expectations from the feature importance scores. While it is common to make the distinction between local scores that focus on individual predictions and global scores that look at the contribution of a feature to the model, another important division distinguishes \emph{model} scenarios, in which the goal is to understand predictions of a given model from \emph{natural} scenarios, in which the goal is to understand a phenomenon such as a disease.  
We develop a set of axioms that represent the properties expected from a feature importance function in the natural scenario and prove that there exists only one function that satisfies all of them, the \emph{\underline{M}arginal \underline{C}ontribution Feature \underline{I}mportance} (MCI). We analyze this function for its theoretical and empirical properties and compare it to other feature importance scores. While our focus is the natural scenario, we suggest that our axiomatic approach could be carried out in other scenarios too.
\end{abstract}
\section{Introduction}\label{Sec:Intro}
In recent years, increasing amounts of data coupled with the rise of highly complex models such as artificial neural networks have led to major advances in our ability to model highly complex relations~\cite{krizhevsky2012imagenet}. These increasingly complex models tend to be harder to interpret~\cite{lipton2018mythos}. This has led to extensive work on interpretability~\cite{molnar2020interpretable}, explainability~\cite{holzinger2019causability}, and more specifically regarding algorithms that assign feature importance~\cite{lundberg2017unified, plumb2018model, ribeiro2016should, shrikumar2017learning, sundararajan2017axiomatic}. Most of the previous works on feature importance have focused on assigning importance scores for predictions of a specific trained model. Methods for assigning feature importance scores are commonly divided into \textit{local} and \textit{global}, where the goal of local scores is to explain how much each feature effects a specific prediction while the goal of the global scores is to explain how much each feature is effecting the model predictions across the entire data distribution. However, there are scenarios where assigning feature importance to explain a phenomena in the real world is also very useful. For example, consider the case of modeling the relations between gene expression to a certain disease. A scientist may be interested in gene importance as a tool for highlighting the genes most related to that disease to prioritize experiments to be done in the lab.

We identify another important dimension that defines different goals for feature importance scores. This dimension differentiates between \emph{model} feature importance scores, which aim to explain a given model, from the \emph{natural} feature importance scores, which aim to explain a phenomenon in the real world such as factors that contribute to heightened risk of a disease. This raises the question whether feature importance scores that were designed to explain a model would work well in natural scenarios. In this work we argue that the scenarios are different and we show examples where methods that were designed to explain a model would fail to explain a natural phenomenon. Chen and his co-authors~\cite{chen2020true} recently demonstrated the difference between explaining a model and explaining the data for a specific use case in the context of Shapley value. In this work we address this difference more broadly.

%The difference between explaining a model and explaining the data was also demonstrated  in a recent paper of Chen and co-authors~\cite{chen2020true} in the context of the use of Shapley value.
%\acomment{Do you think we should add more details in this paragraph? following to Jason comment, I tried to present the Shapely values based method as an example. Do you think we should explain better its connection to feature importance?}We argue that feature importance scores that best suited for the model scenario are not necessarily best suited for the natural scenario. We provide intuition to this claim by showing examples where other methods, such as Shapley values based methods, misbehave when introduced with duplicated features, and demonstrate this example using an experiment.

Most previous work on feature importance  focused on explaining models~\cite{baehrens2009explain, breiman1984classification,  covert2020understanding,lundberg2017unified, plumb2018model, ribeiro2016should, shrikumar2017learning, sundararajan2017axiomatic}. Therefore, in this work we focus on the natural setting and more specifically on the global-natural scenario. To come up with a good feature importance function for this setting we identify few axioms that any feature importance score in the global-natural case should satisfy to have the properties expected of it. We prove that there is only one function that satisfies these axioms, the \emph{\underline{M}arginal \underline{C}ontribution Feature \underline{I}mportance} (MCI). We compare this score to other feature importance scores, both from a theoretical standpoint and from an empirical one. 

The contributions of this paper are the following:
\begin{enumerate}
    \item We define a new important dimension for feature importance scores that differentiate the model  scenario in which the goal is to explain model predictions from the natural scenario in which the goal is to explain a phenomenon in the real world.
    \item We focus on the global-natural scenario and present three simple properties (axioms) that every feature importance score in this scenario should have.
    \item We prove that these axioms uniquely identify the MCI score. 
    \item We analyse the theoretical properties of MCI and show empirically that it is more accurate and robust than other available solutions.
\end{enumerate}

\section{Feature Importance Scenarios}\label{Sec:Scenarios}
It has been noted that the term ``feature importance'' may have different meanings in different scenarios~\cite{ribeiro2016should}. In the introduction we already discussed the common distinction between \emph{local} and \emph{global} scenarios in which feature importance is used, and suggested another distinction between \emph{model} scenarios and \emph{natural} scenarios. To clarify this two dimensional distinction, in the following we list potential use cases for each scenario.

\paragraph{Global-natural \textnormal{(the scientist scenario):}} When studying the association between gnomic loci or gene expression and a disease it is common for a scientist to train models that predict the disease as a tool to study the disease~\cite{danaee2017deep, mckinney2006machine}. Note that in this case models are acting as a surrogate for the natural phenomenon being studied. Here, feature importance is used to rank loci or genes that may have significant influence on the disease. Note that this use case is not limited to genomics. Similar scenarios have been described in medical studies~\cite{fontana2013estimation}, in chemical sciences~\cite{mccloskey2019using}, and other domains~\cite{cui2010music}.

\paragraph{Global-model \textnormal{(the engineer scenario):}} Another common use case for feature importance is in understanding a specific model. For example, for security purposes it is important to understand the sensitivity of a model to changes in specific features~\cite{guo2018lemna}. In other cases, feature importance is used for explaining a model~\cite{adadi2018peeking} or identifying bugs~\cite{tan2018learning}. Note that main users of the importance scores are the engineers who develop the model and would like to verify its correctness.  

\paragraph{Local-model \textnormal{(the user scenario):}} In this scenario, a user of a predictive model is interested in explanation of a specific prediction~\cite{kulesza2015principles}. For example, it may be a customer trying to understand why an application for credit was denied by a credit-provider and what can the customer do to change this decision. Here feature importance serves in understanding how the model used by the credit provider sees the customer.

\paragraph{Local-natural \textnormal{(the patient scenario):}} Consider analyzing the CT scan of a patient for identifying cancer tumors. Important features here are places in the image that are cancerous. In this case, the goal in highlighting important features is to identify the location of the tumor in the body, and the model is only a tool for doing that.
%While the user scenarios and the patient scenarios belong to the family of local scenarios, both engineers scenarios and scientists scenarios are global scenarios. However, in the engineers scenarios the focus is on a specific model while in scientists scenarios the model is a tool to study natural structures. In the same way, in the user scenario the focus is on a specific model while in the patient scenario the model is a proxy for understanding what is happening in the human body. Therefore, we call scenarios in which we are trying to understand a specific model \emph{model scenarios} while scenarios in which the models are tools, or proxies, \emph{natural scenarios}. Making this distinction we can see that user scenarios are local-model scenarios, patient scenarios are local-natural scenarios, engineers scenarios are global-model scenarios while scientists scenarios are global-natural scenarios. 

To summarize, in model scenarios the goal is to understand a specific model whereas in the natural scenarios the model is just a tool for learning a natural structure. Further, in the local scenarios the goal is to understand the effect of each feature on a specific prediction or case, whereas in the global scenario the goal is to understand the global effect of each feature.

This differences also reflect in the expectations of feature importance scores. While the difference between the local and global settings has been discussed before~\cite{ribeiro2016should, tan2018learning, tonekaboni2020went}, we would like to explain here some differences between the model scenarios and the natural scenarios scores. %Imagine for example, that there are two features in the data that are identical. In a model scenario, if the model decided to use just one of them then the feature the model uses is consider important while the other one not. However, from a scientists point of view, if these features are identical then they should have the same importance since they are indistinguishable in the available data.
%\acomment{Do you think that this paragraph can replace the duplication example above?}
Imagine a scientist investigating the relation between gene expression to a disease, while some of those genes are highly correlated in their expression. From a scientist point of view, correlated genes encapsulate the same information about the disease, and therefore they are equally important. On the other hand, from an engineer's point of view, a model built for making predictions using the same data  could use just a portion of those genes for the sake of making accurate predictions. In this case, an engineer trying to understand the model would like to consider only the specific genes the model uses.

From the previous discussion, it follows that feature importance scores that may work well in the model setting will not necessarily be adequate for the natural setting. In the rest of the work, we focus our attention on the global-natural scenario that has received limited attention thus far.

\section{Definitions and Notations}\label{Sec:Prev}
In this section, we introduce important concepts that were defined by previous works~\cite{covert2020understanding} and are relevant to the global-natural scenario. These concepts will serve us in defining our method. We start our discussion with \emph{Shapley value}, a fundamental concept in game theory that was 
recently adopted to the realm of feature importance~\cite{lundberg2017unified, covert2020understanding}. Shapley value was originally designed for problems of cost allocation where $n$ participants cooperate to achieve a certain good~\cite{shapley1953value}. Shapley value uses an \emph{evaluation function} $\nu: 2^n \mapsto \R^+$, that computes for each set of players $S$ the value generated if these participants were to cooperate, where for the empty set the evaluation is assumed to be zero.

%Some feature importance scores are specific to a single type of models. For example, in linear models it is common practice to look at weights assigned to each feature as a measure for its importance while in tree based models a common practice is to look at the sum of the gains from decision nodes in which a feature was used~\cite{breiman1984classification}. MAPLE is a method to explain ensemble of trees~\cite{plumb2018model} while for deep learning there are methods such as Integrated Gradients~\cite{sundararajan2017axiomatic} and DeepLift~\cite{shrikumar2017learning}. Examples for model agnostic methods include the well-known SHAP method of Lundberg and Lee~\cite{lundberg2017unified} or the more recent SAGE method~\cite{covert2020understanding}, both using the Shapley Value~\cite{shapley1953value} as their main building block.

%at first adopted to feature selection~\cite{cohen2007feature} methods, and then extended to the realm of local-model feature scenario by Lundberg and Lee~\cite{lundberg2017unified} with the celebrated SHAP method, and recently to the global-model scenario with SAGE~\cite{covert2020understanding}.

Shapley presented four axioms that a fair allocation of cost should have and showed that there is only one fair cost allocation function~\cite{shapley1953value}:
\begin{equation}
I_\nu\left(f\right)=\frac{1}{\left|F\right|!}\sum_{\sigma\in \pi\left(F\right)}\Delta\left(f,S^{\sigma}_f,\nu\right)\,\,\,.\label{eq:Shapley value}
\end{equation}
where $F$ is the set of all features, $\pi(F)$ is the set of all permutations of $F$, $S^{\sigma}_f$ is the set of all features preceding $f$ in permutation $\sigma$, and 
$
\Delta\left(f,S,\nu\right)=\nu\left(S\cup\left\{ f\right\} \right)-\nu\left(S\right)
$.

By treating features as  players cooperating to make accurate predictions, this idea was adopted for feature selection~\cite{cohen2007feature} and then was extended to local-model feature importance by the SHAP method~\cite{lundberg2017unified}, and recently also extended to the scenario of global-model feature importance by the SAGE method~\cite{covert2020understanding}.

In the context of global feature importance, the evaluation function is a measure of the \emph{predictive power} of a given set of features~\cite{covert2020understanding}. 
For example, let $G(S)$ be the set of predictors restricted to use only the set of features $S$, then given a loss function $\ell$, the predictive power $\nu$ can be defined as:
$$
\nu(S)=\min_{g \in G(\emptyset)}\E_{(x,y)}\left[\ell\left(g(x),y\right)\right]
-
\min_{g \in G(S)}\E_{(x,y)}\left[\ell\left(g(x),y\right)\right] 
$$
where the expectation is with respect to the distribution of the feature vector $x$ and the label $y$.
Other examples of measures of predictive power include the AUC of the best classifier in $G(S)$ or the mutual information between the set of features and the label.\footnote{In all cases we apply additive normalization such that $\nu(\emptyset)=0$.} 

In this work, we refer to a function $\nu:2^F \mapsto \R^+$ as an evaluation function if $\nu(\emptyset)=0$ and if $\nu$ is monotone increasing in the sense that if $S\subseteq T$ then $\nu(S)\leq\nu(T)$. This reflects the intuition that giving more features to the model can only increase the amount of information on the label and thus allows more accurate models.

While the Shapley values axioms make sense in the realm of allocating costs to beneficiaries, we question their adequacy to the natural scenario of feature importance. To demonstrate the problem, consider a system with the binary features $f_1,f_2,f_3$ which are Rademacher random variables such that the target variable $y=f_1 \wedge (f_2 \vee f_3)$ and the mutual information is the evaluation function. Shapley value assigns the feature importance $0.65, 0.15,$ and $0.15$ to $f_1, f_2,$ and $f_3$ respectively. However, if $f_1$ is duplicated 3 times then the feature importance becomes $0.15, 0.18,$ and $0.18$ for $f_1, f_2,$ and $f_3$. Note that the importance of $f_1$ drops when it is duplicated while the importance of $f_2$ and $f_3$ increases to the point that they become the most important features. This means that if these features were indicators of the presence of a certain protein in a blood sample then their importance scores may change if measurements are taken of other proteins that are triggered by the same mechanism and therefore are highly correlated. As a consequence, if the scientist suspected that a certain mechanism is responsible for a disease and therefore sampled many proteins that are related to this mechanism then Shapley-based feature importance scores will suggest that these proteins are of lesser importance.  

\emph{Ablation studies} are another common approach to assign importance scores to features~\cite{bengtson-roth-2008-understanding, casagrande1974ablation,hessel2018rainbow}. In this method, the importance of a feature is the reduction in performance due to removing this feature. Using the notation above, in ablation studies the importance of feature $f$ is $I_\nu(f) = \nu(F) - \nu(F \setminus \{f\})$.

\emph{Bivariate association} is the complement to ablation studies. In this method feature importance is its contribution in isolation, that is, $I_\nu(f) = \nu(\{f\})$. These methods are commonly used in Genome-Wide Association Studies (GWAS)~\cite{haljas2018bivariate, liu2009powerful}, in feature ranking methods~\cite{Zien_2009}, feature selection methods~\cite{guyon2003introduction}, or in feature screening methods~\cite{fan2008sure}.

As mentioned in~\cite{covert2020understanding}, both the ablation and the bivariate association methods deal imperfectly with specific types of feature interactions. Ablation and Shapley-based methods fail when features are highly correlated, while the bivariate method fails when there are synergies between the features. As an example consider an XOR where the target variable is the exclusive or of two features - bivariate methods would fail to find the association between these features and the target variable. %To overcome these difficulties we introduce the MCI score in the next section. 

So far we have discussed importance scores that are model agnostic. However, it is important to mention that there are also importance scores that are specific to a certain type of models. For example, in linear models, it is common practice to look at weights assigned to each feature as a measure for its importance while in tree-based models, a common practice is to look at the sum of the gains from decision nodes in which a feature was used~\cite{breiman1984classification}. MAPLE is a method to explain ensemble of trees~\cite{plumb2018model} while for deep learning there are methods such as Integrated Gradients~\cite{sundararajan2017axiomatic} and DeepLift~\cite{shrikumar2017learning}.

We now move forward to introducing our method, which aims to overcome the difficulties in existing methods.  

\section{Marginal Contribution Feature Importance}\label{Sec:Method}
In previous sections, we discussed the different scenarios in which feature importance can be used and presented the limitations of current methods in the global-natural scenario. To find a proper score for this scenario we begin by introducing a small set of properties expected of a feature importance scoring function in this setting which we refer to as axioms. We show that Marginal Contribution Feature Importance (MCI) is the only function that satisfies these axioms and we study its properties.

\subsection{The Axioms}
The axioms use the \emph{Elimination} operation, which is defined as follows:
\begin{definition}
Let $F$ be a set of features and $\nu$ be an evaluation function. Eliminating  the set $T\subset F$  creates a new set of features $F^{\prime}=F\setminus T$ and a new valuation function $\nu^{ \prime }:F^\prime\mapsto \R^{+}$ such that $\forall S\subseteq F^{\prime},\,\,\,\nu^{\prime}\left(S\right)=\nu\left(S\right)$.
\end{definition}

Using this definition we introduce the set of required properties (axioms):

\begin{definition}\label{def:axioms}
A valid feature importance function $I_{\nu}$ in the global-natural scenario is a function $I_{\nu}:F\mapsto \R^{+}$ that has the following properties:
\begin{enumerate}
    \item \textbf{Marginal contribution}: The importance of a feature is equal or higher than the increase in the evaluation function when adding it to all the other features, i.e $I_\nu(f) \geq (v(F) - v(F/\{f\}))$.\label{axiom:marginal contribution}
    \item \textbf{Elimination}: Eliminating  features from $F$ can only decrease the importance of each feature. \emph{i.e.}, if\: $T \subseteq F$ and $\nu^\prime$ is the evaluation function which is obtained by eliminating $T$ from $F$ then $\forall f \in F\setminus T$,~~~ $I_\nu(f) \geq I_{\nu^\prime}(f)$.\label{axiom:elimination}
    \item \textbf{Minimalism}: If $I_{\nu}$ is the feature importance function, then for every function $I:F\mapsto\R^+$ for which axioms~\ref{axiom:marginal contribution} and~\ref{axiom:elimination} hold, and for every $f\in F$: $I_\nu(f)\leq I(f)$.
\end{enumerate}
\end{definition}

To understand the rationale behind these axioms, the \emph{Marginal contribution} states that if a feature generates an increase of performance even when all other features are present, then its importance is at least as large as this additional gain it creates. This is to say that if the ablation study (see section~\ref{Sec:Prev}) shows a certain gain, then the feature importance is at least this gain.

The rationale for the \emph{Elimination} axiom is that the importance of a feature may be apparent only when some context is present. For example, if the target variable is the XOR of two features, their significance is apparent only when both of them are observed. Therefore, eliminating features can cause the estimated importance of the remaining features to drop. On the other hand, if a feature shown to be important, that is, it provides high predictive power given the current set of features, then its predictive power does not decrease when additional information is provided by adding features. Note that it still may be the case that the relative importance of features changes when adding or eliminating features. In other words, we may find, by adding features, that a feature that was considered less important is very significant in combination with these new features. However, if a feature was considered having high predictive power then adding feature can only demonstrate even higher predictive power using the additional information provided by the new featues.

%\acomment{My concern about this explanation is that we show an example where > might hold, but we don't argue that $\geq$ is a must, or that $<$ can't hold}To understand the \emph{Elimination}  property consider first the XOR problem: if the target variable is the XOR over a set of features then we expect the feature importance function to give a positive importance score to all variables participating in the XOR function. However, if we eliminate some features that participate in the XOR function we may no longer be able to uncover the relation between the features and the target variables and may, therefore, conclude these features to be less important. This property is not limited to the XOR setting: whenever there is synergy between features  eliminating features may decrease the observed importance of a feature due to missing some of the context required to see its significance.

Finally, note that if $I_{\nu}$ satisfies the marginal contribution and the elimination properties, then for every $\lambda > 1$ the function $\lambda I_{\nu}$ also satisfies these properties. The \emph{Minimalism} axiom provide disambiguation by requiring the selection of the smallest function.   

These axioms allow us to present the main theorem which shows the existence and uniqueness of the feature importance function.
\begin{theorem}\label{THM: exists and unique} The function 
$$I_{\nu}\left(f\right)=\max\limits_{S\subseteq F}\Delta\left(f,S,\nu\right)$$ 
is the only function that satisfies the marginal contribution, the elimination and the minimalism axioms. 
\end{theorem}
 
 Theorem~\ref{THM: exists and unique} shows that there is only one way to define a feature importance function that satisfies the axioms presented above. We call this function, the Marginal Contribution Feature Importance (MCI) function. The proof of this theorem is presented in the supplementary material (Section~A).

\subsection{Properties of the Marginal Contribution Feature Importance Function}\label{Sec:Properties}
MCI has many advantageous properties as shown in the following theorem.

\begin{theorem}\label{THM: similarity to shapley values} Let $F$ be a set of features, let $\nu$ be an evaluation
function and let $I_{\nu}$ be the feature importance
function $I_{\nu}\left(f\right)=\max\limits_{S\subseteq F}\Delta\left(f,S,\nu\right)$.
The following holds:
\begin{itemize}
\item \textbf{Dummy}: if $f$ is a dummy variable, that is $\forall S\subseteq F,\,\,\,\Delta\left(f,S,\nu\right)=0$,
then $I_{\nu}\left(f\right)=0$.
\item \textbf{Symmetry}: if $f_{i}$ and $f_{j}$ are symmetric, that is
if for every $S\subseteq F$
we have that $\nu\left(S\cup\left\{ f_{i}\right\} \right)=\nu\left(S\cup\left\{ f_{j}\right\} \right)$,
then $I_{\nu}\left(f_{i}\right)=I_{\nu}\left(f_{j}\right)$.
\item \textbf{Super-efficiency}: $\forall S\subseteq F,\,\,\,\nu\left(S\right)\leq\sum_{f\in S}I_{\nu}\left(f\right)$.
\item \textbf{Sub-additivity}: if $\nu$ and $\omega$ are evaluation functions
defined on $F$ then $\forall f\in F,\,\,\,I_{\nu+\omega}\left(f\right)\leq I_{\nu}\left(f\right)+I_{\omega}\left(f\right)$.
    \item \textbf{Upper bound the self contribution}: for every feature $f \in F$, $I_\nu(f) \geq \nu(\{f\})$.
    \item \textbf{Duplication invariance}: let $F$ be a set of features and $\nu$ be an evaluation function. Assume that $f_{i}$ is a duplication of $f_{j}$ in the sense that for every $S\subseteq F\setminus\left\{ f_{i},f_{j}\right\} $ we have that
$\nu\left(S\cup\left\{ f_{i},f_{j}\right\} \right)=\nu\left(S\cup\left\{ f_{i}\right\} \right)=\nu\left(S\cup\left\{ f_{j}\right\} \right)$.
If $F^{\prime}$ and $\nu^{\prime}$ are the results of eliminating
$f_{i}$ then $\forall f\in F^{\prime},\,\,\,I_{\nu^{\prime}}\left(f\right)=I_{\nu}\left(f\right)$.
\end{itemize}
\end{theorem}

Recall that the Shapley value is defined by four axioms: efficiency,
symmetry, dummy and additivity~\cite{shapley1953value}. Theorem~\ref{THM: similarity to shapley values} shows that MCI has the symmetry and dummy properties, but the efficiency
property is replaced by a super-efficiency property while the additivity property is replaced by a sub-additivity property. The upper bound on self contribution shows that MCI always dominate the bivariate association scores. It is also easy to verify that it upper bounds Shapley value and the ablation score. 
Finally, duplication invariance shows that when features are duplicated the feature importance scores do not change, this is unlike Shapley value for which we showed earlier its sensitivity to duplication. 
The proof of Theorem~\ref{THM: similarity to shapley values} is presented in the supplementary material (Section~A). 

Another interesting property of MCI is the context it can provide for the importance of a feature. From the definition of MCI it follows that for every $f$ there is $S\subseteq F$ such that $I_\nu(f)=\Delta\left(f,S,\nu\right)$. This $S$ is a context with which $f$ provides a big gain. In some cases this context can give additional insight to the scientist.

\subsection{Computation and Approximation}
The complexity of computing the  MCI function in a straight forward way is exponential in the number of features. Since computing the Shapley value is NP-complete~\cite{deng1994complexity}, there is no reason to believe that MCI is easier to compute. In the supplementary material (Section~C) we provide examples for cases were MCI can be computed in polynomial time, for example, when $\nu$ is sub-modular. Moreover, much like Shapley value, MCI can be approximated by sampling techniques~\cite{castro2009polynomial}. One interesting property of MCI is that any sampling based technique provides a lower-bound on the score. We also present some upper-bounds that allow for a branch and bound type technique. In addition, we show that when there are strong correlations between features an estimation for MCI scores can be derived from computing the scores on a much smaller problem.

Another challenge in computing MCI is obtaining the values of $\nu$ for various sets. Using the theory of uniform convergence of empirical means (the PAC model)~\cite{vapnik2015uniform}, we show that with high probability $\nu$ can be estimated to within an additive factor using a finite sample and this estimate can be used to estimate MCI to within a similar additive factor. Details about computation and approximation techniques are provided in the supporting material (Section~C).

\section{Experiments}\label{Sec:experiments}
In this section we analyze the performance of MCI empirically and compare it to other methods. For the experiments described here we used the
breast cancer sub-type classification task from a gene microarray dataset (BRCA)~\cite{tomczak2015cancer}. We present two experiments, in the first one we compare the quality of the ranking of the feature importance provided by different methods. In the second experiment we test the robustness of the methods to correlations between the features. In the supplementary material (Section~E) we also provide results for robustness experiments on six different datasets from the UCI repository~\cite{asuncion2007uci}.
%We first describe the data that was used. Then, we provide details on the experiments' design and implementation, including a proposed approximation algorithm. %that was used to deal with the high computational complexity of the problem in hand.
%and to deal with the challenge of approximating the evaluation function using a finite sample.
%Next, we present the experiments' results and show that the proposed method yields better importance ranking quality.%successful in giving the highest rank to the genes that are known to be most relevant to breast cancer. 
%Also, we demonstrate how other methods are affected significantly by adding features duplicates, while our method remains stable.

\subsection{Data}
We used a BRCA~\cite{tomczak2015cancer} dataset that consists of 17,814 genes from 571 patients that have been diagnosed for one of 4 breast cancer sub-types. In our experiments we used the same subset of 50 features used in~\cite{covert2020understanding}. The main advantage of this dataset is the existing scientific knowledge about genes that are related to breast cancer. We used this knowledge to evaluate the quality of different feature importance scores. 

%In order to estimate MCI using a finite sample we need to deal with two challenges: estimate $\nu$ for any $S \subseteq F$, and given the estimation of $\nu$ we need to deal with the exponential time complexity of MCI.
\subsection{Implementation Details}
The evaluation of any features subset $S \subseteq F$ was conducted by defining $\nu(S)$ to be the average negative log-loss over 3-fold cross validations of a logistic regression model trained using only the features in $S$. 

Since computing Shapley values and MCI scores exactly has exponential complexity, we used the sampling technique proposed by~\cite{covert2020understanding} for the SAGE algorithm. According to this method, a random set of permutations $P_d$ of the features is sampled. For each $\sigma\in P_d$ we denote by $S_i^\sigma = \{f_j : \sigma(j)<\sigma(i)\}$ and estimate the feature importance for Shapley and MCI to be:
$$
I_\nu^{\mbox{Shapley}}(f_i) = \frac{1}{\left\vert P_d\right\vert}\sum_{\sigma\in P_d}\Delta(f_i,S_i^\sigma,\nu)~~~~\mbox{and}~~~~I_\nu^{\mbox{MCI}}(f_i) = \max_{\sigma\in P_d}\Delta(f_i,S_i^\sigma,\nu)~~~~.
$$
Recall that using this method provides an unbiased estimator for the Shapley value and a lower bound for MCI. In our experiments with BRCA $P_d$ was of the size $2^{15}$ as we observed that the relative order of features stabilizes in both methods at this point. %For the sake of completeness, we also ran SAGE~\cite{covert2020understanding} even though this method is addressing the global-model scenario and not the global-natural scenario.
\begin{table}[t]
  \caption{\bf{Results of experiments with the BRCA dataset}:  \textmd{On the left hand side the results for Experiment I, measuring the quality of the scores, are presented using the NDCG scores of prefixes of varying sizes of the rankings generated by the different methods (higher is better, the perfect score is $1.00$). On the right side the results for Experiment II, measuring the robustness of the scores to feature correlations. The results are the Minimal Kendall Distance (MKD) between the rankings of the top-$k$ ranked elements  before and after duplicating the top feature three times. In these results lower is better, the perfect score is 0.00.}}
  \label{tab:BRCA Exp.I and Exp.II}
  \label{sample-table}
  \centering
  \begin{tabular}{lcccccccccc}
    \toprule
    &
    \multicolumn{5}{c}{\bf{Experiment I: Quality }} &
    \multicolumn{5}{c}{\bf{Experiment II: Robustness }} \\
    & 
    \multicolumn{5}{c}{NDCG $\uparrow$} &
    \multicolumn{5}{c}{MKD $\downarrow$} \\
%    \cmidrule(r){2-6}
%    \cmidrule(r){7-11}\\
    Method& @3& @5 & @10 & @20 & @50 & @3& @5 & @10 & @20 & @50 \\
    \cmidrule(r){1-1}
    \cmidrule(r){2-6}
    \cmidrule(r){7-11}
    MCI &  \textbf{1.00} &  \textbf{0.85} &  \textbf{0.77} &  \textbf{0.88} &  \textbf{0.92} &  \textbf{0.00} &  \textbf{0.00} &  0.04 &  0.02 &  0.03 \\
    Shapley &  0.77 &  0.70 &  0.73 &  \textbf{0.88} &  0.88  &  1.00 &  0.28 &  0.13 &  0.08 &  0.04 \\
    Bivariate &  \textbf{1.00} &  \textbf{0.85} &  \textbf{0.77} &  0.82 &  \textbf{0.92} & \textbf{0.00} &  \textbf{0.00} &  \textbf{0.00} &  \textbf{0.00} &  \textbf{0.00} \\
    Ablation              &  0.30 &  0.21 &  0.28 &  0.44 & 0.61 &  0.12 &  0.14 &  0.17 &  0.09 &  0.04 \\
    \bottomrule
  \end{tabular}
\end{table}

\subsection{Experiment I: Quality}
%\textbf{Experiment's results.}
The goal of the first experiment is to evaluate the quality of the importance scores provided by the different methods. Each feature importance score generates a ranking of the features. Since there is existing scientific knowledge about the association between genes and breast cancer sub-types, we consider a better ranking to be one that gives a higher ranking to genes that are known to be related to breast cancer~\cite{covert2020understanding}. 
We evaluated the ranking using the well-known Normalized Discounted Cumulative Gain (NDCG) metric~\cite{jarvelin2002cumulated}. 
%This metric was originally designed to evaluate the performance of search engine's documents ranking given the relevance grade of each document.
%Similarly, assuming that genes that were previously found to be related to different breast cancer subtypes could be used as this pre-defined knowledge and should be considered as more important for predicting these subtypes, we can use this a-priori knowledge to evaluated each feature importance ranking. 
The results of the experiment are presented in Table~\ref{tab:BRCA Exp.I and Exp.II}. The results show that MCI and Bivariate outperform Shapley while Ablation performs poorly. The success of the Bivariate method in this experiment suggests that there are no significant synergies between the features in this dataset. MCI handled this situation and even outperforms Bivariate slightly in the top 20 list. However, due to the strong correlations between features, Ablation failed to generate a meaningful ranking and this is a probable explanation also to the low performance of Shapley.

SAGE~\cite{covert2020understanding} is a global feature importance score that was designed for the global-model scenario. It is possible to use it in the global-natural scenario by training a model on a  portion of the data and using the rest for evaluating feature importance. Unfortunately, we found that when used in this manner, SAGE's results are sensitive to the choice of the random seed (other methods were found to be stable) and in most cases it is substantially outperformed by MCI. Details are provided in the supplementary material (Section~F). 

%In addition to the global-natural methods, we also evaluated the ranking quality of the SAGE method, which address the global-model scenario. Due to high variance in SAGE results for runs with different random seeds, we struggled with evaluating its performance. In the supplementary material we show results of each of the methods for 5 different random seeds. We show that all the methods except from SAGE were highly stable for different random seeds, and that MCI outperforms SAGE in 4 out of 5 runs. 
 %In addition, since in this case adding a single gene to all the others is barley improving the model performance, we can see that the Ablation study based method preformed purely with respect to the others.\\

\begin{figure*}[t!]
    \centering
    \begin{subfigure}[t]{0.5\textwidth}
        \centering
        \includegraphics[height=1.3in]{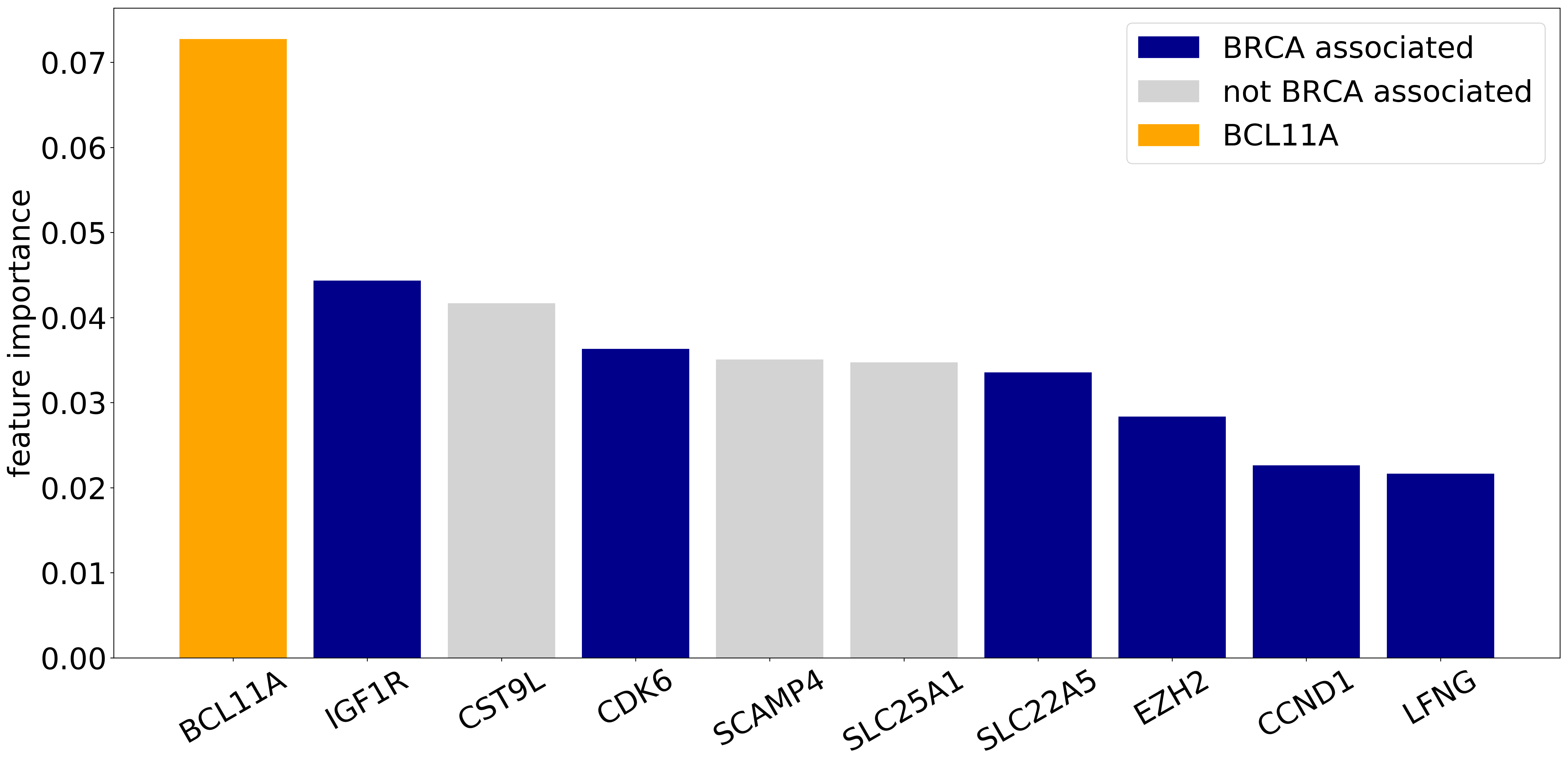}
        \caption{Shapley values}
    \end{subfigure}%
    ~
    \begin{subfigure}[t]{0.5\textwidth}
        \centering
        \includegraphics[height=1.3in]{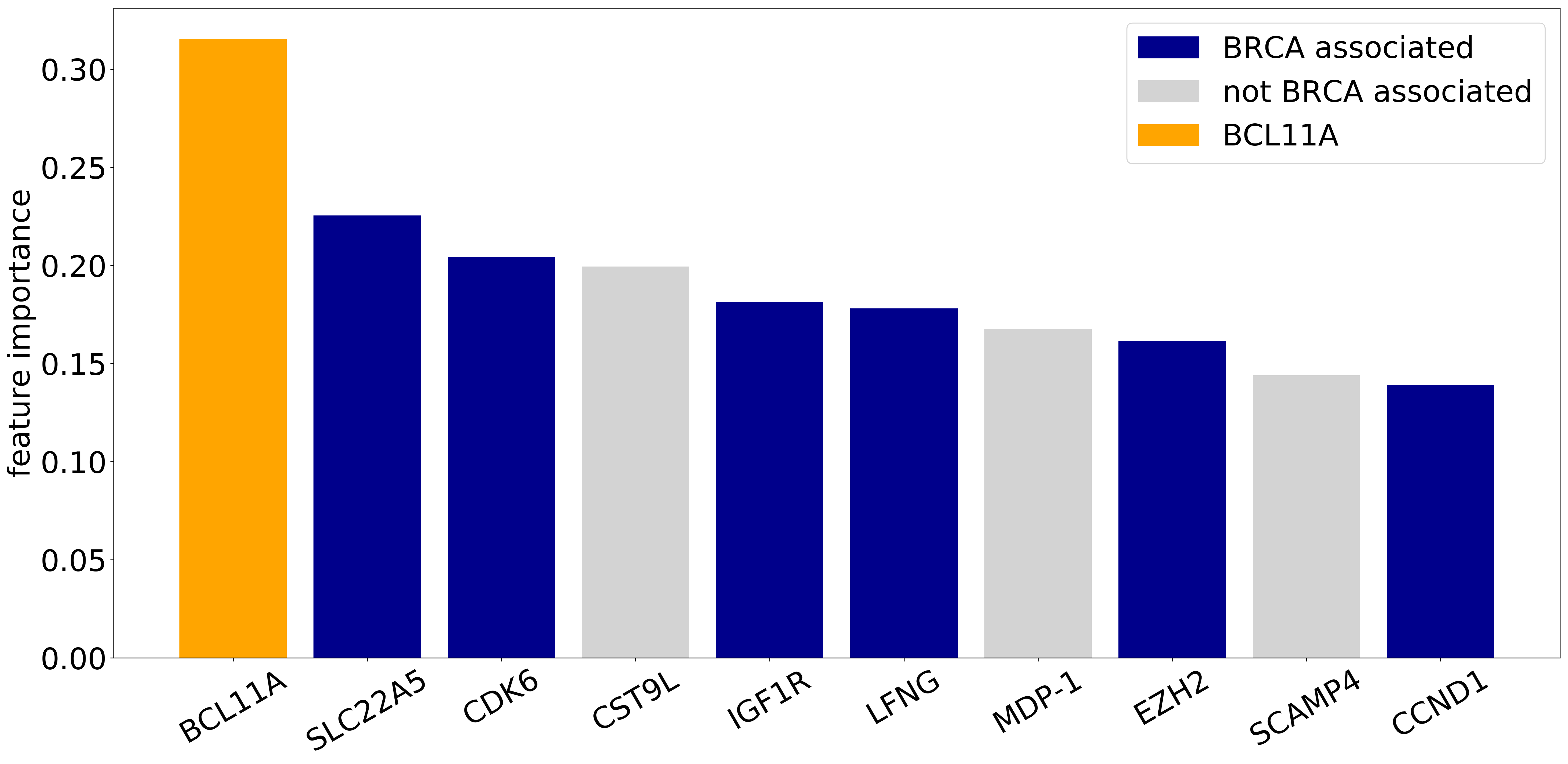}
        \caption{MCI values}
    \end{subfigure}
    \\
    \begin{subfigure}[t]{0.5\textwidth}
        \centering
        \includegraphics[height=1.3in]{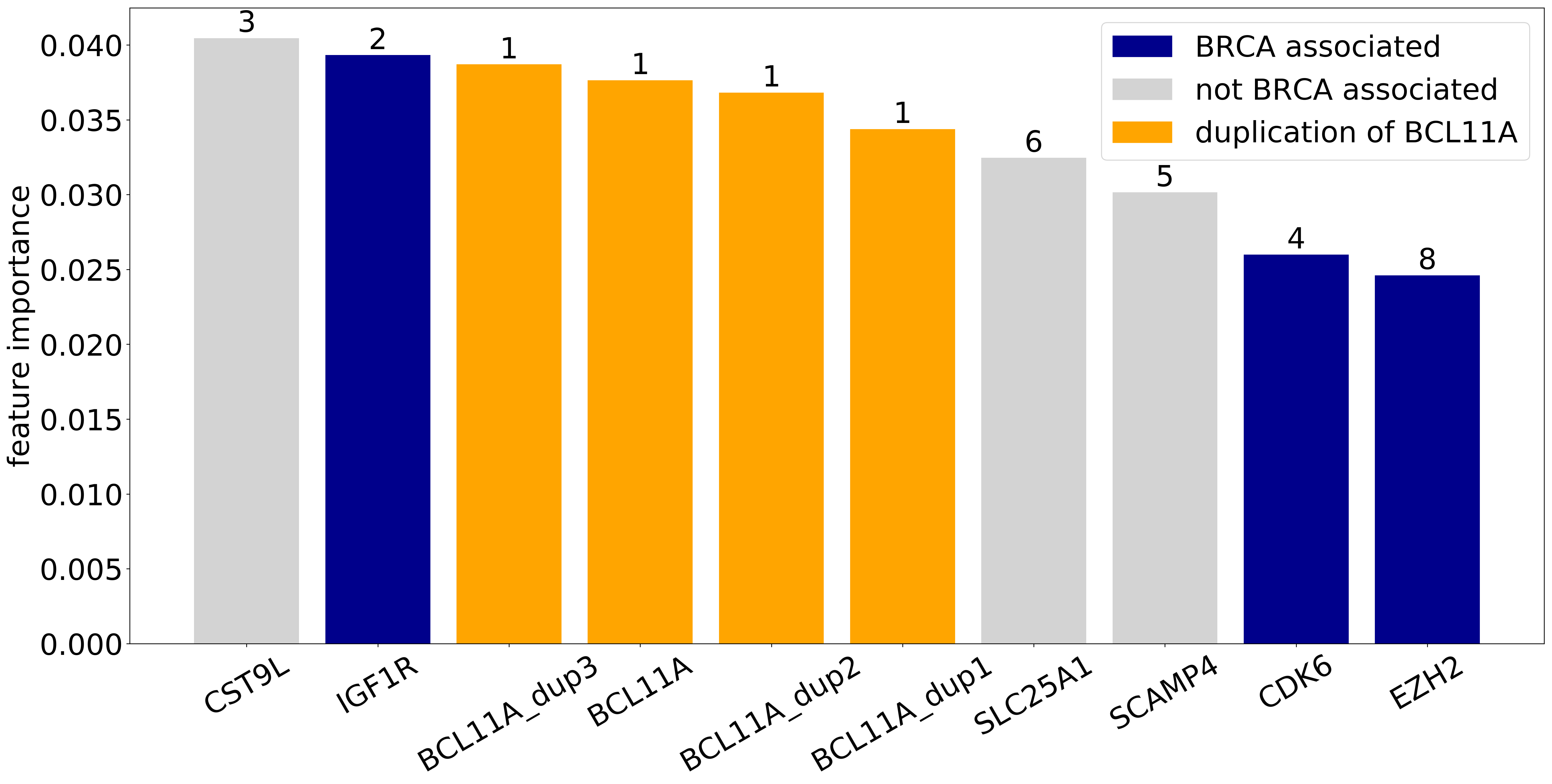}
        \caption{Shapley values with duplication}
    \end{subfigure}%
    ~
    \begin{subfigure}[t]{0.5\textwidth}
        \centering
        \includegraphics[height=1.3in]{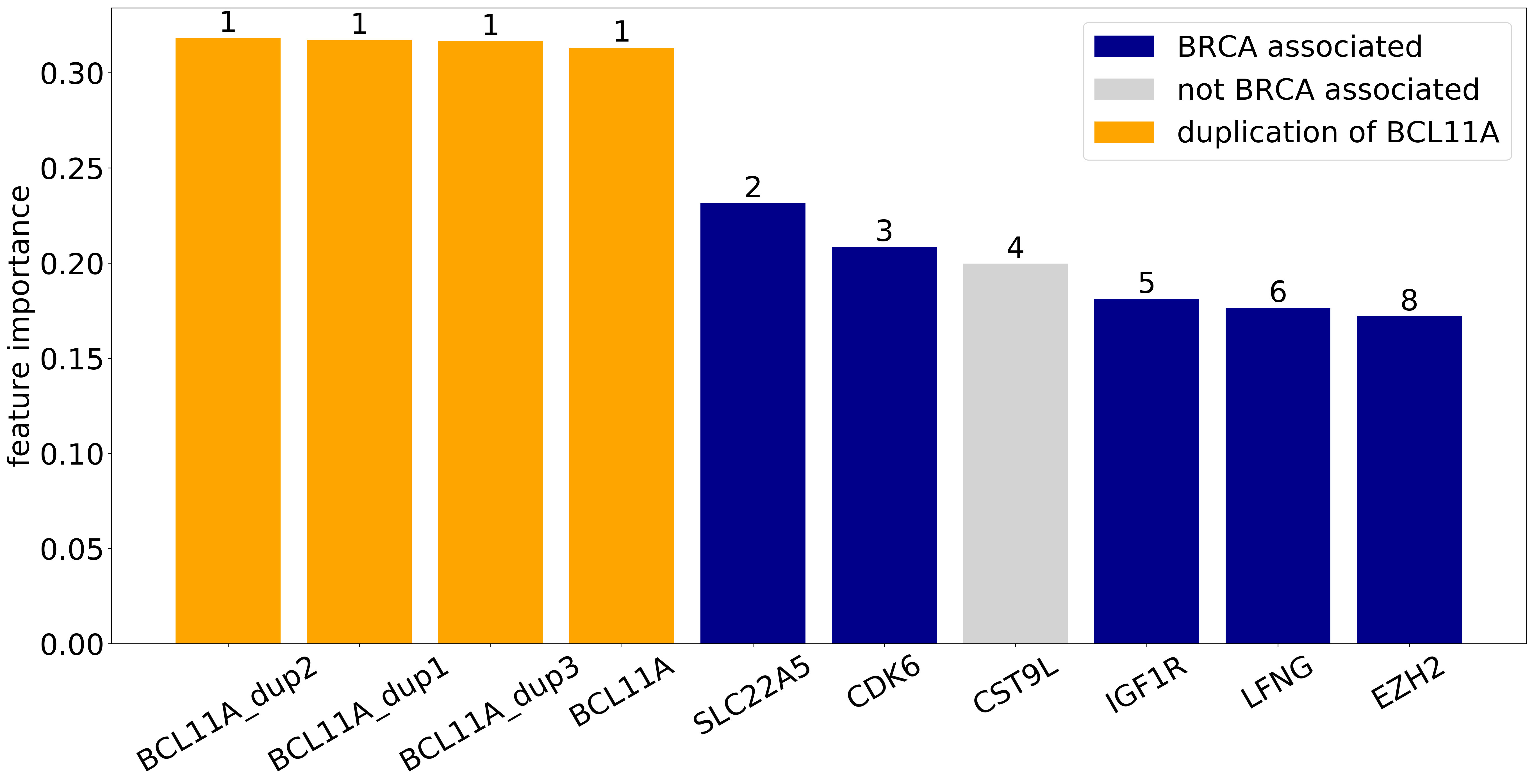}
        \caption{MCI values with duplication}
    \end{subfigure}
    \caption{Feature duplication experiment on the BRCA dataset, with the top 10 ranking features displayed. On the top row we show the feature importance according to Shapley (a) and MCI (b). The bottom row shows the estimations of both methods, when the top ranked feature (BCL11A) is duplicated three time. Both methods produce similar feature importance estimation when the top ranked feature is not duplicated. However, when it is, Shapley' importance assignment (c) is affected drastically, while MCI (d) succeeds to remain stable.}
    \label{fig:dup_experiment}
\end{figure*}

%In table~\ref{table:no_dups}, we can see that MCI outperformed all the other methods in the sense of ranking the breast cancer related genes higher. In addition, since in this case adding a single gene to all the others is barley improving the model performance, we can see that the Ablation study based method preformed purely with respect to the others.
\subsection{Experiment II: Robustness}
The goal of the second experiment, is to measure the robustness of the different methods and more specifically the implications of correlated features on the ranking they produce. For each method we duplicated three times the feature that was ranked first and re-evaluated the scores and the associated ranking with the additional features. We computed the distance between the top-$k$ list before duplicating the feature to the top-$k$ list after the duplication.

To measure the distances between the rankings of the top-$k$ features we used the Minimizing Kendall Distance (MKD)~\cite{fagin2003comparing} metric. In this method, for the ranking $r_k$ of the top-$k$ features we define $C(r_k)$ to be the set of completions of $r_k$ to rankings of the entire feature set. \ie, $r\in C(r_k)$ if $r$ is a ranking of all the features and $\forall i=1,\ldots,k,~~~r(i)=r_k(i)$. Using this notation the MKD distance between $r_k$ and $r_k^\prime$ is $\mkd(r_k,r_k^\prime)=\min_{r\in C(r_k), r^\prime\in C(r_k^\prime )} d_\tau(r,r^\prime)$ where $d_\tau$ is the Kendall-tau distance. Furthermore, since we introduced  duplicated features which are not ranked in the original ranking, we included only the first occurrence of this feature in the list of ranked features.

The results of this experiment are presented in Table~\ref{tab:BRCA Exp.I and Exp.II}. As expected, the Bivariate method is not effected by the duplicate features since it ignores feature correlations. MCI was resilient to the introduction of these features in the top of the list while some small changes are seen at the bottom of the list. We suspect that this is noise due to the fact that only a small sample was used to evaluate $\nu$. Shapley suffered the most from the introduction of new feature and most of the differences are at the top of the list. \ie, features that were considered the most important features before the duplication were pushed down the list. Finally, Ablation suffered from the introduction of duplicated features too.

Additional demonstration of the difference between MCI and Shapley is presented in Figure~\ref{fig:dup_experiment}. In this figure the top 10 features computed by Shapley and MCI before and after introducing the duplicated features are presented. While the rankings before the duplication are similar in both methods, this is no longer the case after the duplication. While MCI finds all the duplicates to be as important as the original feature, Shapley penalizes for the duplication and demotes this feature while promoting the feature that was on the $3^{\mbox{rd}}$ place to  the $1^{\mbox{st}}$ place.

\section{Conclusion}\label{Sec:Conclusion}
In this work we introduced an important distinction between the model scenario of feature importance scores, and the natural scenario. We showed in what sense these scenarios are different, and demonstrated cases where previous methods, such as Shapley values based methods, behave imperfectly when extended to the natural scenario. Therefore, we focused on the natural scenario and used an axiomatic approach in which we identified three properties any feature importance function for this scenario should satisfy. We showed that there exists a function that satisfy all of these properties, and that it is unique. We called it the MCI function and addressed its computational and approximation challenges in the supplementary material. Finally, we ran several experiments showing that the MCI method performed better in identifying important features, and was robust to addition of correlated features, whereas Shapley values based methods are highly affected from this kind of change.
%\subimport{./}{impact}

\appendix

\renewcommand\thesection{\Alph{section}}
\renewcommand\thesubsection{\thesection.\Roman{subsection}}
\section{Proofs}\label{Sec:proofs}
In this section we group together all the proofs for the theorems we presented in the main paper.

\subsection{Existence and Uniqueness of the Feature Importance Score}
Here we prove Theorem~\ref{THM: exists and unique} states that there is a function for which all the axioms defined in Definition~\ref{def:axioms} hold and that this function is unique. This proof is a constructive proof in the sense that we are able to show that the only feature importance function for which the axioms hold is the function $I_\nu$ that assigns to the feature $f\in F$ the score
\[I_v(f) = \max_{S\subseteq{F}} (v(S \cup \{f\}) - v(S)) = \max_{S\subseteq F}\Delta\left(f,S,\nu\right)~~~.\]
Meaning, the importance of a feature is the maximum contribution to the valuation function $\nu$ over any subset of features. 

\begin{lemma} \label{LEMMA: lower bound}
Let $I_{\nu}$ be a feature importance function for which axioms (1), (2) hold (marginal contribution, and elimination axiom). Then:
\[
I_{\nu}\left(f\right)\geq\max_{S\subseteq F}\Delta\left(f,S,\nu\right)\,\,\,.
\]
\end{lemma}

\begin{proof}
We prove the statement using induction on the size of the features set $F$. Let $n=\left|F\right|$. If $n=1$ (i.e $F = \{f\}$) then from the marginal contribution axiom we have that

\begin{eqnarray*}
I_{\nu}\left(f\right) & \geq & \nu\left(\left\{ f\right\} \right)-\nu\left(\emptyset\right)\\
 & = & \max_{S\subseteq F}\Delta\left(f,S,\nu\right)\,\,\,.
\end{eqnarray*}

%As it holds that $\Delta(f, \emptyset, \nu) \geq \Delta(f, \{f\}, \nu) = 0$ since $\nu$ is a function to $\R^+$ and $\nu(\emptyset) = 0$. \\

Assume that the statement holds for any set of features of size $< n$ for $n>1$. Let
$\left|F\right|=n$ and let $f\in F$. Let $\Star=\arg\max\limits_{S\subseteq F}\Delta\left(f,S,\nu\right)$.
If there exists $f^{\prime}\in F\setminus\left\{ \Star\cup\left\{ f\right\} \right\} $
then from the elimination axiom, if $\left\{ f^{\prime}\right\} $
is eliminated we will obtain $F^{\prime}$ and $\nu^{\prime}$ such
that $\left|F^{\prime}\right|=n-1$ and $I_{\nu^{\prime}}\left(f\right)\leq I_{\nu}\left(f\right)$.
However, since $\Star\subseteq F^{\prime}$ we have that from the assumption
of the induction
\[
I_{\nu}\left(f\right)\geq I_{\nu^{\prime}}\left(f\right)\geq\Delta\left(f,\Star,\nu^{\prime}\right)=\Delta\left(f,\Star,\nu\right)=\max_{S\subseteq F}\Delta\left(f,S,\nu\right)\,\,\,.
\]
Otherwise, assume that $\Star=\arg\max\limits_{S\subseteq F}\Delta\left(f,S,\nu\right)$
is such that $\Star\cup\left\{ f\right\}=F$. Therefore, $\max\limits_{S\subseteq F}\Delta\left(f,S,\nu\right)=\Delta\left(f,F,\nu\right)$.
From the marginal contribution axiom we have that $I_{\nu}\left(f\right)\geq\Delta\left(f,F,\nu\right)=\max\limits_{S\subseteq F}\Delta\left(f,S,\nu\right)$.\\
\end{proof}

Lemma~\ref{LEMMA: lower bound} shows that any importance function that has the marginal contribution property and the eliminate property must assign an importance score of at least $\max\limits_{S\subseteq F} \Delta\left(f,S,\nu\right)$ to every feature. Therefore, by adding the minimalism axiom we obtain the uniqueness and existence of the feature importance function as shown in Theorem~\ref{THM: exists and unique}.

Here we prove Theorem~\ref{THM: exists and unique}
\begin{proof} Adding the minimalism axiom to Lemma~\ref{LEMMA: lower bound}
shows that if the marginal contribution and the elimination axioms hold for $I_{\nu}\left(f\right)=\max\limits_{S\subseteq F}\Delta\left(f,S,\nu\right)$
then it is the unique feature importance function. Proving that the marginal contribution axiom hold is straight-forward: for a feature $f\in F$
\[
I_{\nu}\left(f\right)=\max\limits_{S\subseteq F}\Delta\left(f,S,\nu\right)\geq\Delta\left(f,F \setminus \{f\},\nu\right)=\nu\left(F\right)-\nu\left(F\setminus\{f\}\right)~~~.
\]
To see that the elimination axiom holds too, let $T\subset F$ and let $f\in F\setminus T$. If $T$ is eliminated from $F$ to create $F^{\prime}$ and $\nu^{\prime}$ then
\[
I_{\nu}\left(f\right)=\max_{S\subseteq F}\Delta\left(f,S,\nu\right)\geq\max_{S\subseteq F^{\prime}}\Delta\left(f,S,\nu\right)=\max_{S\subseteq F^{\prime}}\Delta\left(f,S,\nu^{\prime}\right)=I_{\nu^{\prime}}\left(f\right)\,\,\,.
\]
\end{proof}

\subsection{Properties of the MCI Function}
Here, we prove the MCI function properties presented in Theorem~\ref{THM: similarity to shapley values}.
\begin{proof} 
\textbf{Dummy}:  Let $f$ be a dummy variable such that $\forall S\subseteq F,\,\,\,\Delta\left(f,S,\nu\right)=0$
then $I_{\nu}\left(f\right)=\max\limits_{S\subseteq F}\Delta\left(f,S,\nu\right)=0$.

\textbf{Symmetry}: Let $f_{i}$ and $f_{j}$ be such that for every $S\subseteq F$ we have that
$\nu\left(S\cup\left\{ f_{i}\right\} \right)=\nu\left(S\cup\left\{ f_{j}\right\} \right)$. Consider any set $S\subseteq F$. We consider three cases, (1) $f_i, f_j \in S$, (2) $f_i,f_j \notin S$, and (3) exactly one of $f_i,f_j$ is in $S$. In case (1) we have that $\Delta\left(f_i,S,\nu\right)=\Delta\left(f_j,S,\nu\right)=0$. In case (2) we have
$$
\Delta\left(f_i,S,\nu\right)=\nu\left(S\cup\{f_i\}\right)-\nu\left(S\right) = \nu\left(S\cup\{f_j\}\right)-\nu\left(S\right)=\Delta\left(f_j,S,\nu\right)~~~.
$$
In case (3) assume, w.l.o.g. that $f_i\in S$ and $f_j \notin S$. Let $S^\prime$ denote the set $S$ where $f_i$ is replaced by $f_j$ and therefore, due to the symmetry between $f_i$ and $f_j$ it holds that $\nu\left(S\right)=\nu\left(S^\prime\right)$. Note also that $S\cup\{f_j\}=S^\prime\cup\{f_i\}$ and therefore $\Delta\left(f_i,S,\nu\right)=\Delta\left(f_j,S^\prime,\nu\right)$. 
From analyzing these 3 cases it follows that for every $S\subseteq F$ there exists $S^\prime \subseteq F$ such that $\Delta\left(f_i,S,\nu\right)=\Delta\left(f_j,S^\prime,\nu\right)$. Therefore, $I_\nu(f_i)\leq I_\nu(f_j)$. However, by replacing the roles of $f_i$ and $f_j$ it also holds that $I_\nu(f_i)\geq I_\nu(f_j)$ and therefore $I_\nu(f_i)= I_\nu(f_j)$.

\textbf{Super-efficiency}: Let $S\subseteq F$. w.l.o.g. let $S=\left\{ f_{1},f_{2},\ldots,f_{k}\right\} $.
Define $\forall_{i \leq k} S_{i}=\left\{ f_{j}\right\} _{j\leq i}$. Therefore, $S_{0}=\emptyset$
and $S_{k}=S$. Since $\nu(\emptyset) = 0$,
\begin{eqnarray*}
\nu\left(S\right)  &=&  \nu\left(S_{k}\right)-\nu\left(S_{0}\right)
  =  \sum_{i=0}^{k-1}\left(\nu\left(S_{i+1}\right)-\nu\left(S_{i}\right)\right)\\
  &=&  \sum_{i=0}^{k-1}\Delta\left(f_{i+1},S_{i},\nu\right)
  \leq  \sum_{i=0}^{k-1}I_{\nu}\left(f_{i+1}\right)
  =  \sum_{f\in S}I_{\nu}\left(f\right)\,\,\,.
\end{eqnarray*}

\textbf{Sub-additivity}: if $\nu$ and $\omega$ are valuation functions
defined on $F$ then for all $f\in F$
\begin{eqnarray*}
I_{\nu+\omega}\left(f\right)  &=&  \max_{S\subseteq F}\Delta\left(f,S,v+\omega\right)
  =  \max_{S\subseteq F}\left(\Delta\left(f,S,\nu\right)+\Delta\left(f,S,\omega\right)\right)\\
  &\leq&  \max_{S\subseteq F}\Delta\left(f,S,\nu\right)+\max_{S\subseteq F}\Delta\left(f,S,\omega\right)
  =  I_{\nu}\left(f\right)+I_{\omega}\left(f\right)\,\,\,.
\end{eqnarray*}

\textbf{Upper bound the self contribution:} For $f \in F$ We have that $I_{\nu}\left(f\right)=\max\limits_{S\subseteq F}\Delta\left(f,S,\nu\right)\geq \Delta\left(f,\emptyset,\nu\right)=\nu\left(\{f\}\right)-\nu\left(\emptyset\right)=\nu\left(\{f\}\right)$.

\textbf{Duplication invariant:} Assume that $f_{i} \in F$ is a duplication of $f_{j} \in F$
in the sense that for every
$S\subseteq F\setminus\left\{ f_{i},f_{j}\right\} $ we have that
$\nu\left(S\cup\left\{ f_{i},f_{j}\right\} \right)=\nu\left(S\cup\left\{ f_{i}\right\} \right)=\nu\left(S\cup\left\{ f_{j}\right\} \right)$.
Let $F^{\prime}$ and $\nu^{\prime}$ be the results of eliminating
$\{f_{i}\}$ and let $f\in F^{\prime}$. From the elimination axiom we have that $I_\nu\left(f\right)\geq I_{\nu^\prime}\left(f\right)$. Assume that $\Star\subseteq F$ is such that $I_\nu (f)=\Delta(f,\Star,\nu)$. If $f_i\notin \Star$ then $\Star \subseteq F^\prime$ and $I_{\nu^\prime}\left(f\right)\geq \Delta(f,\Star,\nu^\prime) = \Delta(f,\Star,\nu) = I_{\nu}\left(f\right)$. Otherwise $f_i \in \Star$ and it holds  for $S^\prime = \Star \cup\{f_j\}\setminus\{f_i\}$ that $\Delta(f,\Star,\nu)=\Delta(f,S^\prime,\nu)=\Delta(f,S^\prime,\nu^\prime)\leq I_{\nu^\prime}(f)$. Therefore, in all possible cases we have that $I_\nu\left(f\right)\leq I_{\nu^\prime}\left(f\right)$. When combined with the elimination axiom we conclude that $I_\nu\left(f\right)= I_{\nu^\prime}\left(f\right)$.
\end{proof}

\section{Additional Properties of the MCI Function}\label{Sec:properties}

In the following theorem we present and prove additional relevant properties of the MCI function, that did not mentioned in the main paper. 

\begin{theorem}\label{THM: additional properties}
Let $F$ be a set of features, let $\nu$ be a valuation function and let $I_{\nu}$ be the MCI function. 
The following holds:
\begin{itemize}
\item \textbf{Scaling}:  $\forall f\in F,\,\,\,\forall \lambda>0,\,\,\,\,I_{\lambda\nu}\left(f\right)=\lambda I_{\nu}\left(f\right)$.
\item \textbf{Monotonicity}: If  $\forall S \subseteq F\setminus\{f_i,f_j\},\,\,v\left(S\cup \{f_{i}\}\right)\leq v\left(S\cup \{f_{j}\}\right)$
then $I_{v}\left(f_{i}\right)\leq I_{v}\left(f_{j}\right)$.
\end{itemize}
\end{theorem}

In the following we prove theorem~\ref{THM: additional properties}
\begin{proof}

\textbf{Scaling:} This property follows since for every $f$ and every $S$ it holds that $\lambda\Delta(f,S,\nu) = \Delta(f,S,\lambda\nu)$.

\textbf{Monotonicity:} Let $f_i, f_j \in F$ for which $\forall S \subseteq F\setminus\{f_i,f_j\},\,\,v\left(S\cup \{f_{i}\}\right)\leq v\left(S\cup \{f_{j}\}\right)$. Let $\Star$ be such that $I_{\nu}\left(f_{i}\right)=\Delta(f_i,\Star,\nu)$. Then, if $f_{j}\notin \Star$ it holds that 
$$
I_{\nu}\left(f_{i}\right)=\Delta(f_i,\Star,\nu)\leq \Delta(f_j,\Star,\nu)\leq I_{\nu}\left(f_{j}\right) ~~~.
$$
Otherwise, if $f_j\in \Star$ then
$$
I_{\nu}\left(f_{i}\right)=\Delta(f_i,\Star,\nu)\leq \Delta(f_j,\Star\cup \{f_i\}\setminus\{f_j\},\nu)\leq I_{\nu}\left(f_{j}\right) ~~~.
$$

\end{proof}
\section{Computation Optimizations}
 We now turn our attention to the computational challenge of computing or approximating the MCI function. Straight-forward computation is exponential in the size of the feature set. Therefore we study cases in which the computation can be made efficient and approximation techniques.

\subsection{Submodularity}
In the following we show that if the valuation function $\nu$ is submodular~\cite{lovasz1983submodular} then the MCI feature importance score is equal to the self contribution of each feature.

\begin{lemma}\label{LEMMA: sub-modularity}
If $\nu$ is sub-modular then $I_{\nu}\left(f\right)=\nu\left(\left\{ f\right\} \right)$.
\end{lemma}

\begin{proof} Recall that in that case there is a diminishing return
and therefore for every $S\subseteq F\setminus\left\{ f\right\} $:$\nu\left(S\cup\left\{ f\right\} \right)\leq\nu\left(S\right)+\nu\left(\left\{ f\right\} \right)$. Therefore, 
$$
\Delta(f,S,\nu)=\nu(S\cup\{f\})-\nu(S)\leq \nu(\{f\})=\Delta(f,\emptyset,\nu)
$$
\end{proof}

The submodularity assumption might be too stringent in some cases. For example, if the target variable is an XOR of some features then the submodularity assumption does not hold. However, if we assume that submodularity holds for large sets then we obtain a polynomial algorithm for computing the feature importance. This may make sense in the genomics setting where genes may have synergies but we may assume that only small interactions of 2,3, or 4 genes are significant. We begin by defining $k$-size submodularity:
\begin{definition}\label{DEF: k-size-submodularity}
A function $\nu:F\mapsto\R$ is $k$-size submodular if for every $S,T\subseteq F$ such that $|S|,|T|\geq k$ 
$$
\nu(S)+\nu(T)\geq \nu(S\cup T) + \nu(S\cap T)
$$
A function $\nu:F\mapsto\R$ is soft $k$-size submodular if it holds  that for every $T\subseteq F$, $|T| > k$, $f \in F$ there exists $S \subseteq T$, $|S| \leq k$ for which:
$$
\nu(S \cup{\{f\}})+\nu(T)\geq \nu((S\cup{\{f\}}) \cup T) + \nu((S\cup{\{f\}})\cap T)
$$
\end{definition}
\begin{lemma}\label{LEMMA: k-size-sub-modularity}
If $\nu$ is $k$-size-submodular or soft $k$-size-submodular then
$$
I_\nu(f)=\max_{S\subseteq F~:~|S|\leq k}\Delta(f,S,\nu)
$$
\end{lemma}
\begin{proof}
First, we show that if $\nu$ is $k$-size-submodular it is also soft $k$-size-submodular. Let $\nu$ be a $k$-size-submodular valuation function. Let $T \subseteq F$, $|T| > k$ and let $S \subseteq T$, $|S| \geq k$, $f \in F$. From the $k$-size-submodular property we get that:
$$
\nu(S \cup{\{f\}})+\nu(T)\geq \nu((S\cup{\{f\}}) \cup T) + \nu((S\cup{\{f\}})\cap T)
$$
And therefore $\nu$ is also soft $k$-size-submodular. Hence, it is enough to prove the theorem for soft $k$-size-submodular functions.\\
Let $\nu$ be a soft $k$-size-submodular evaluation function. Let $T\subseteq{F}$ be such that $T=\arg\min\left\{|T|~:~I_\nu(f)=\Delta(f,T,\nu) \right\}$.  Assume, in contradiction, that $|T|>k$. Note that $f\notin T$ since if $f\in T$ then $I_\nu(f)=\Delta(f,T,\nu)=0$, and in this case we have that $I_\nu(f)=\Delta(f,\emptyset,\nu)$ in contradiction. Due to the soft $k$-size submodular and monotonous properties of $\nu$ it follows that exists $S\subseteq{T}$, $|S| \leq k$ such that:
$$
\nu(S\cup\{f\})+\nu(T)\geq \nu(T\cup\{f\})+\nu(S)
$$
Therefore,
$\Delta(f,S,\nu)=\nu(S\cup\{f\})-\nu(S)\geq \nu(T\cup\{f\})-\nu(T)=\Delta(f,T,\nu)$. This is a contradiction since $|S|<|T|$.
\end{proof}
Lemma~\ref{LEMMA: k-size-sub-modularity} shows that if $\nu$ is  soft $k$-size-submodular then the entire function $I_\nu$ can be computed in time
$O\left(|F|^{k+1}\right)$.

\subsection{Branch and Bound Optimization}
Here we show
how we can discard computation for some of the subsets using a branch and bound like technique.

\begin{lemma} For every $S_{0}\subseteq S_{1}\subseteq S_{2}\subseteq F$
and $f\in F$:
\[
\Delta\left(f,S_{1},\nu\right)\leq\nu\left(S_{2}\cup \{f\}\right)-\nu\left(S_{0}\right)
\]

\end{lemma}

\begin{proof} 
\label{Lemma:upper-bound}This lemma follows from the monotonous property of $\nu$.
$$
\Delta\left(f,S_{1},\nu\right)  =  \nu\left(S_{1}\cup \{f\}\right)-\nu\left(S_{1}\right)
  \leq  \nu\left(S_{2}\cup \{f\}\right)-\nu\left(S_{0}\right)~~~.
$$\end{proof}

The ability to upper bound $I_\nu$ provided by this Lemma allows  cutting back the computation significantly. For example, if we computed $\Delta(f,S,\nu)$ for every set $S$ of size $k$ and we have that $\maxl{S: \vert S \vert \leq k} \Delta(f,S,\nu) \geq \maxl{S: \vert S \vert = k} \nu(F)-\nu(S)$ then $I_\nu(f)=\maxl{S: \vert S \vert \leq k} \Delta(f,S,\nu)$. The following lemma proves this property in a more general setting.

\begin{lemma}
  Let $\Sbb = \{S\subseteq F \mbox{ s.t. } \vert S \vert = K\}$  and $\Tbb = \{T\subseteq F \mbox{ s.t. } \vert T \vert = k\}$ for $0\leq k \leq K\leq \vert F \vert$.   Let $\bar{\Sbb}=\{S\subseteq F: \exists S^\prime \in \Sbb ~\mbox{s.t.}~ S^\prime \subseteq S \}$  and $\underline{\Tbb}=\{T\subseteq F: \exists T^\prime \in \Tbb ~\mbox{s.t.}~ T \subseteq T^\prime \}$. Let $s_f=\maxl{S\in\bar{\Sbb}}\Delta(f,S,\nu)$ and  $t_f=\maxl{T\in\underline{\Tbb}}\Delta(f,T,\nu)$  and $st_f  = \maxl{S\in\Sbb, T \in \Tbb} \nu(S\cup{f})-\nu(T)$ then 
  $$
  \max (s_f,t_f) \leq I_\nu(f) \leq \max(s_f,t_f, st_f)~~~.
  $$
\end{lemma}
\begin{proof}
The lower bound on $I_\nu(f)$ follows from the simple fact that for every $\Sbb\subseteq 2^F$ 
$$
\max_{S\in\Sbb} \Delta(f,S,\nu)\leq I_\nu(f)~~~.
$$
Let $\Star$  be such that $I_\nu(f)=\Delta(f,\Star,\nu)$. If $\Star \in \bar{\Sbb} \cup \underline{\Tbb}$ then $I_\nu(f)=\max(s_t,t_f)$. Otherwise, there exists $S\in\Sbb$  and $T\in \Tbb$ such that $T\subset \Star \subset S$  and from Lemma~\ref{Lemma:upper-bound} it holds that 
$$
I_\nu(f)=\Delta(f,\Star,\nu)\leq\nu(S\cup{f})-\nu(T)
$$
which completes the proof.
\end{proof}
\subsection{Heuristics}
Recall that for any $\Sbb\subseteq 2^F$  it holds that $\maxl{S\in\Sbb} \Delta(f,S,\nu)\leq I_\nu(f)$. Therefore, any method can be used to select $\Sbb$ and obtain a lower bound on the feature importance. In the experiments in this paper we used random permutations to generate the set $\Sbb$  following the proposal of~\cite{covert2020understanding}. This method is described in Section~\ref{Sec:experiments}. Our experiments show that this method is effective. however, in some cases it may be too demanding since for every subset of features a model has to be trained. The computational cost can be further reduced by using a method such as SAGE~\cite{covert2020understanding} to estimate $\Delta(f,S,\nu)$ from a model that was trained on the entire dataset and therefore trained only once. Only for sets $S$ such that SAGE estimates that $\Delta(f,S,\nu)$ is large, the real value can be computed via training models. Therefore, the estimator SAGE (or in any other proposed method) is used to eliminate testing sets $S$ for which the marginal contribution of $f$ is predicted to be small. 
\section{Approximations of the MCI Funciton}

Another challenge in computing $I_\nu$ is computing the evaluation function $\nu$. Recall that we assumed, for example, that $\nu$ is monotone increasing. This is motivated by thinking about $\nu(S)$ as some measure of the information that $S$ provides on the target variable. However, if only a finite sample exists for evaluating $\nu$ then forcing this property is no longer trivial. Therefore, in this section we show that using uniform convergence of empirical means~\cite{vapnik2015uniform}, or in other words, the PAC theory, it is possible to show that $\nu$ can be approximated well enough from a finite sample such that the estimate of $I_\nu$ will be good too.

First, we need to define a way for estimating the valuation function using a finite sample. Let $X$ be a random variable consists of features $F$ and let $Y$ be a target random variable. Let $D$ be a finite i.i.d sample from $X$ and let $\mathcal{H}_S$ be a hypothesis class for any subset of features $S \subseteq F$. For any hypothesis $h \in \mathcal{H}_S$ denote by $e_D(h)$ the expected error of $h$ over $D$, and denote by $e_p(h)$ the expected error of $h$ over the true distribution $p(x,y)$. The estimated valuation function $\nu_D: \mathcal{P}(F) \longrightarrow \R^+$ defined as follows:
$$
\nu_D(S) = \min_{h \in \mathcal{H}_S}\left(e_D\left(h\right)\right)
$$
and the true valuation function $\nu: \mathcal{P}(F) \longrightarrow \R^+$ defined as:
$$
\forall_{S\subseteq F} \: \nu(S) = \min_{h \in \mathcal{H}_S}\left(e_P\left(h\right)\right)
$$

For convenience we assume that $\mathcal{H}_S$ is finite (this restriction can be exchanged for finite VC dimension). denote $\mathcal{|H|} = \max\limits_{S \subseteq F}\left(|\mathcal{H}_S|\right)$. 
\begin{theorem}
Let $X$ be a random variable consists of features $F$, let $Y$ be a target random variable and let $\mathcal{H}_S$ be a hypothesis class for for any subset $S \subseteq F$. \\
For any $\epsilon, \delta > 0$ and sample $D$ of size $m \geq \dfrac{2}{\epsilon^2}\left(\log_2{\left(\dfrac{2|\mathcal{H}|}{\delta}\right)} + |F|\right)$ it holds that:
$$
P\left[\max_{f \in F}|I_{\nu_D}(f) - I_\nu(f)| > \epsilon\right] \leq \delta 
$$
\end{theorem}

\begin{proof}
Let $X$ be a random variable consists of features $F$, let $Y$ be a target random variable and let $\mathcal{H}_S$ be a hypothesis class for any subset of features $S \subseteq F$. Let $\epsilon, \delta > 0$ and let $D$ be an i.i.d sample of size  $m \geq \dfrac{2}{\epsilon^2}\left(\log_2{\left(\dfrac{2|\mathcal{H}|}{\delta}\right)} + |F|\right)$.

First, we show that for all $S \subseteq F$:
\begin{equation*} 
\left|\nu_D(S) - \nu(S)\right| 
 \leq \max_{h \in \mathcal{H}_S}\left|\left(e_D(h) - e_P(h)\right)\right|
\end{equation*}

Let $S \subseteq F$. Denote $h^{*}_{D} = \arg\min_{h \in \mathcal{H}_S}e_D(h)$ and $h^{*} = \arg\min_{h \in \mathcal{H}_S}e_P(h)$. We have that:
$$
\nu_D(S) - \nu(S) 
= e_D(h^{*}_{D}) - e_P(h^{*}) 
\leq e_D(h^{*}) - e_P(h^{*}) 
\leq \max_{h \in \mathcal{H}_S}\left|e_D(h) - e_P(h)\right|
$$
and also:
$$
\nu(S) - \nu_D(S) 
= e_P(h^{*}) - e_D(h^{*}_{D}) 
\leq e_P(h^{*}_D) - e_D(h^{*}_D) 
\leq \max_{h \in \mathcal{H}_S}\left|e_D(h) - e_P(h)\right|
$$

Next we would like to show that for any $f\in F$ it holds that $ 
\left|I_{\nu_D}(f) - I_\nu\right(f)| 
 \leq 2\max_{S \subseteq F}\left|\nu_D(S) - \nu(S)\right|
$.

Let $f \in F$ and let $\Star = \arg\max_{S \subseteq F}\left(\Delta\left(f,S,\nu\right)\right)$, $\Star_D = \arg\max_{S \subseteq F}\left(\Delta\left(f,S,\nu_D\right)\right)$.

Notice that
\begin{eqnarray*} 
I_{\nu_D}(f) - I_\nu(f) & = &
\Delta\left(f,\Star_D,\nu_D\right) - \Delta\left(f,\Star,\nu\right) \\
& \leq & \Delta\left(f,\Star_D,\nu_D\right) - \Delta\left(f,\Star_D,\nu\right) \\
& \leq & \max_{S \subseteq F}\left|\Delta\left(f,S,\nu_D\right) - \Delta\left(f,S,\nu\right)\right|
\end{eqnarray*}

and also
\begin{eqnarray*} 
I_{\nu}(f) - I_{\nu_D}(f) & = &
\Delta\left(f,\Star,\nu\right) - \Delta\left(f,\Star_D,\nu_D\right) \\
& \leq & \Delta\left(f,\Star,\nu\right) - \Delta\left(f,\Star,\nu_D\right) \\
& \leq & \max_{S \subseteq F}\left|\Delta\left(f,S,\nu_D\right) - \Delta\left(f,S,\nu\right)\right|
\end{eqnarray*}

and therefore we get that
\begin{eqnarray*} 
\left|I_{\nu_D}(f) - I_\nu\right(f)| & \leq &  
 \max_{S \subseteq F}\left|\Delta\left(f,S,\nu_D\right) - \Delta\left(f,S,\nu\right)\right| \\
& = & \max_{S \subseteq F}\left| \left(\nu_D(S \cup \{f\}) - \nu_D(S)\right) - \left(\nu(S \cup \{f\}) - \nu(S)\right)  \right| \\
& \leq & \max_{S \subseteq F}\left| \nu_D(S \cup \{f\}) - \nu(S \cup \{f\})\right| + \max_{S \subseteq F}\left|\nu_D(S) - \nu(S)  \right| \\
& \leq & 2\max_{S \subseteq F}\left|\nu_D(S) - \nu(S)\right| 
\end{eqnarray*}

Hence, using union bound and Hoeffding inequality, for any $f \subseteq F$ it holds that:
\begin{eqnarray*}
P\left[\left|I_{\nu_D}(f) - I_\nu\right(f)| > \epsilon\right]
&\leq& P\left[2\max_{S \subseteq F, h \in \mathcal{H}_S}\left|\nu_D(S) - \nu(S)\right| >\epsilon\right] \\
&\leq& P\left[2\max_{S \subseteq F, h \in \mathcal{H}_S}|e_D(h) - e_P(h)| > \epsilon\right]\\
&=& P\left[\bigcup_{S \subseteq F, h \in \mathcal{H}_S} \left\{2|e_D(h) - e_P(h)| > \epsilon\right\}\right]\\
&\leq& \sum_{S \subseteq F, h \in \mathcal{H}_S} P\left[2|e_D(h) - e_P(h)| > \epsilon\right]\\
&\leq& 2^{|F| + 1}|\mathcal{H}|e^{\frac{-n\epsilon^2}{2}} \leq \delta~~~.
\end{eqnarray*}
Where the last inequality follows by using the bound on $m$ in the statement of this theorem.
\end{proof}

\section{Robustness Experiments on UCI Datasets}\label{Sec:UCI}
In this section we present additional experiments comparing the robustness of different feature importance methods. We follow similar procedure to the one  described in Section~\ref{Sec:experiments} for additional six datasets from the UCI repository~\cite{asuncion2007uci} (see description of the datasets in Table~\ref{tab:uci data}). We do not hold quality experiments for these datasets since unlike the BRCA dataset, for these datasets there is no definitive knowledge about the importance of features to which we can compare.

Following the robustness experiment for the BRCA dataset described in Section~\ref{Sec:experiments}, we first computed feature importance using different methods. Then, for each method we duplicated three times the feature that was ranked first and re-compute the feature importance. We measure stability using the MKD distance described in Section~\ref{Sec:experiments}. We computed the MKD distances between the top-$k$ list before the introduction of the duplicates, and the top-$k$ list after the duplication.

% \usepackage{booktabs}
\begin{table}[tbh]
\caption{\bf{Description of the UCI datasets used in the robustness experiment}}\label{tab:uci data}
\centering
\begin{tabular}{@{}llcc@{}}
\toprule
\textbf{Dataset}               & \textbf{Type}  & \multicolumn{1}{c}{\textbf{\# Features}} & \multicolumn{1}{c}{\textbf{\# Examples}} \\ \midrule
\textbf{Heart Disease}         & Classification & 13                                       & 303                                      \\
\textbf{Wine Quality}          & Regression     & 11                                       & 1599                                     \\
\textbf{German Credit Default} & Classification & 20                                       & 1000                                     \\ 
\textbf{Bike Rental}           & Regression     & 12                                       & 303                                      \\
\textbf{Online Shopping}       & Classification & 17                                       & 12330                                    \\
\textbf{Bank Marketing}        & Classification & 16                                       & 45211                              \\
\bottomrule
\end{tabular}
\end{table}

\begin{figure*}[p]
    \centering
    \begin{subfigure}[t]{0.5\textwidth}
        \centering
        \includegraphics[height=1.3in]{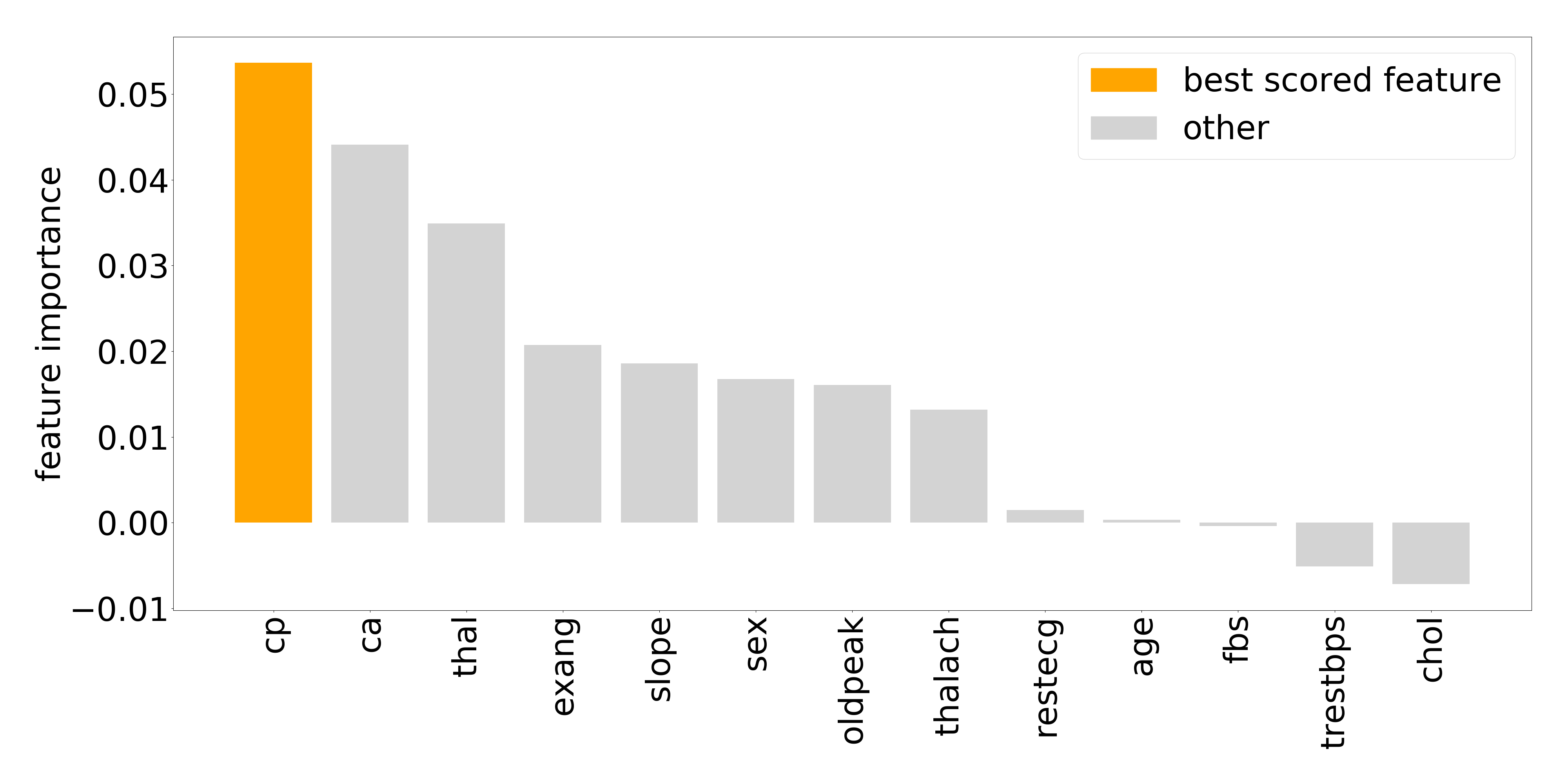}
        \caption{Shapley values}
    \end{subfigure}%
    ~
    \begin{subfigure}[t]{0.5\textwidth}
        \centering
        \includegraphics[height=1.3in]{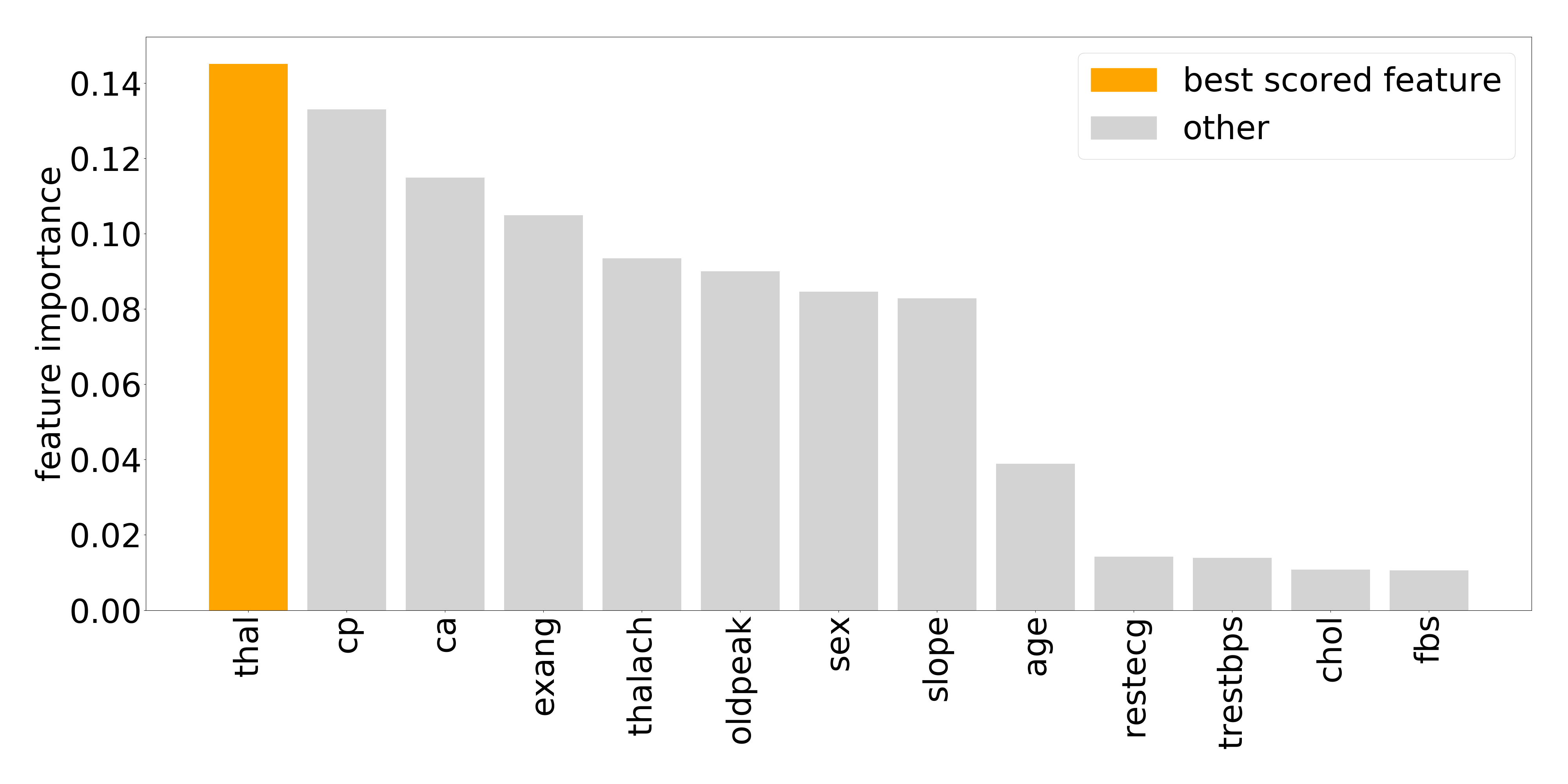}
        \caption{MCI values}
    \end{subfigure}
    \\
    \begin{subfigure}[t]{0.5\textwidth}
        \centering
        \includegraphics[height=1.3in]{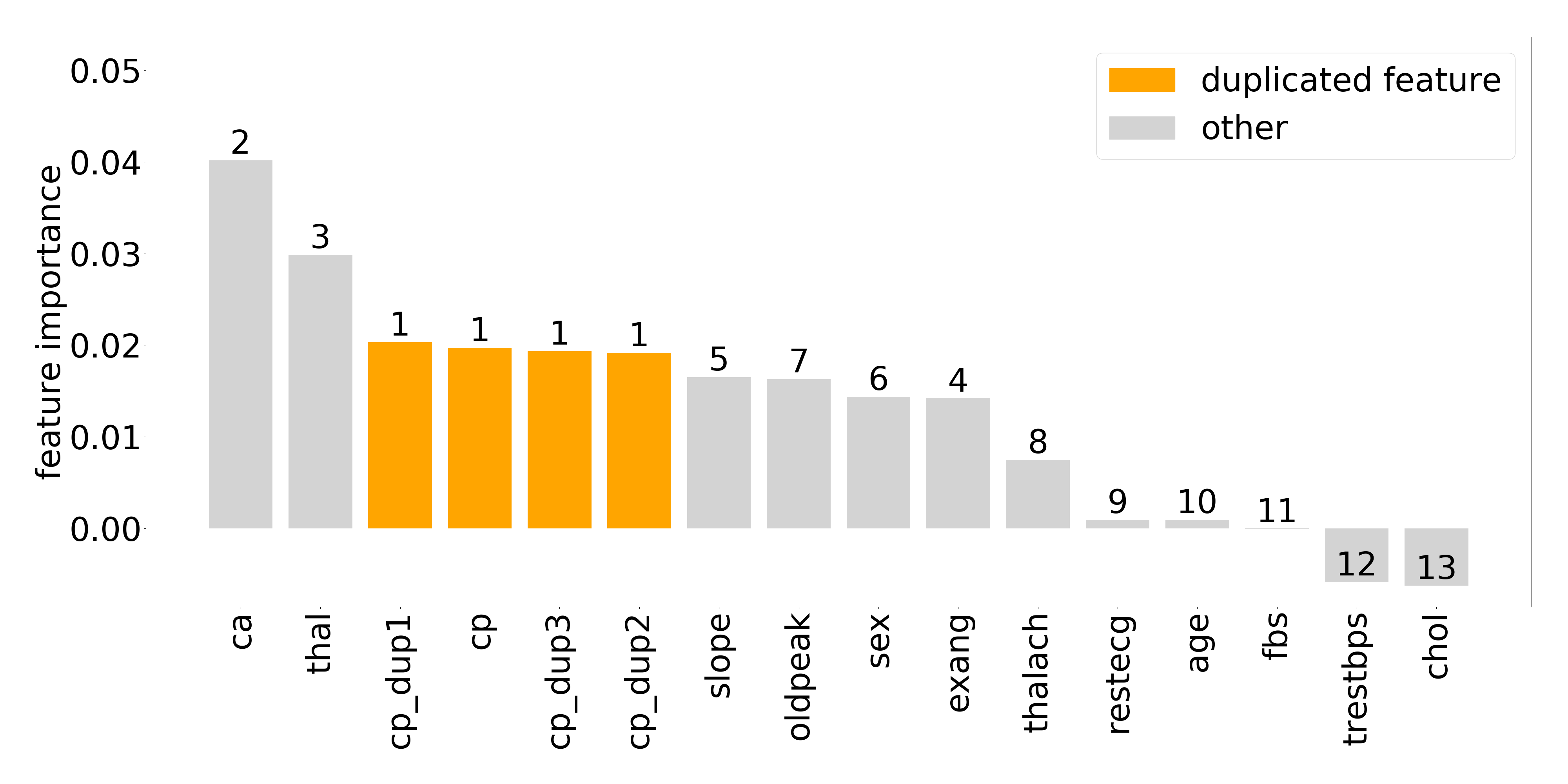}
        \caption{Shapley values with duplication}
    \end{subfigure}%
    ~
    \begin{subfigure}[t]{0.5\textwidth}
        \centering
        \includegraphics[height=1.3in]{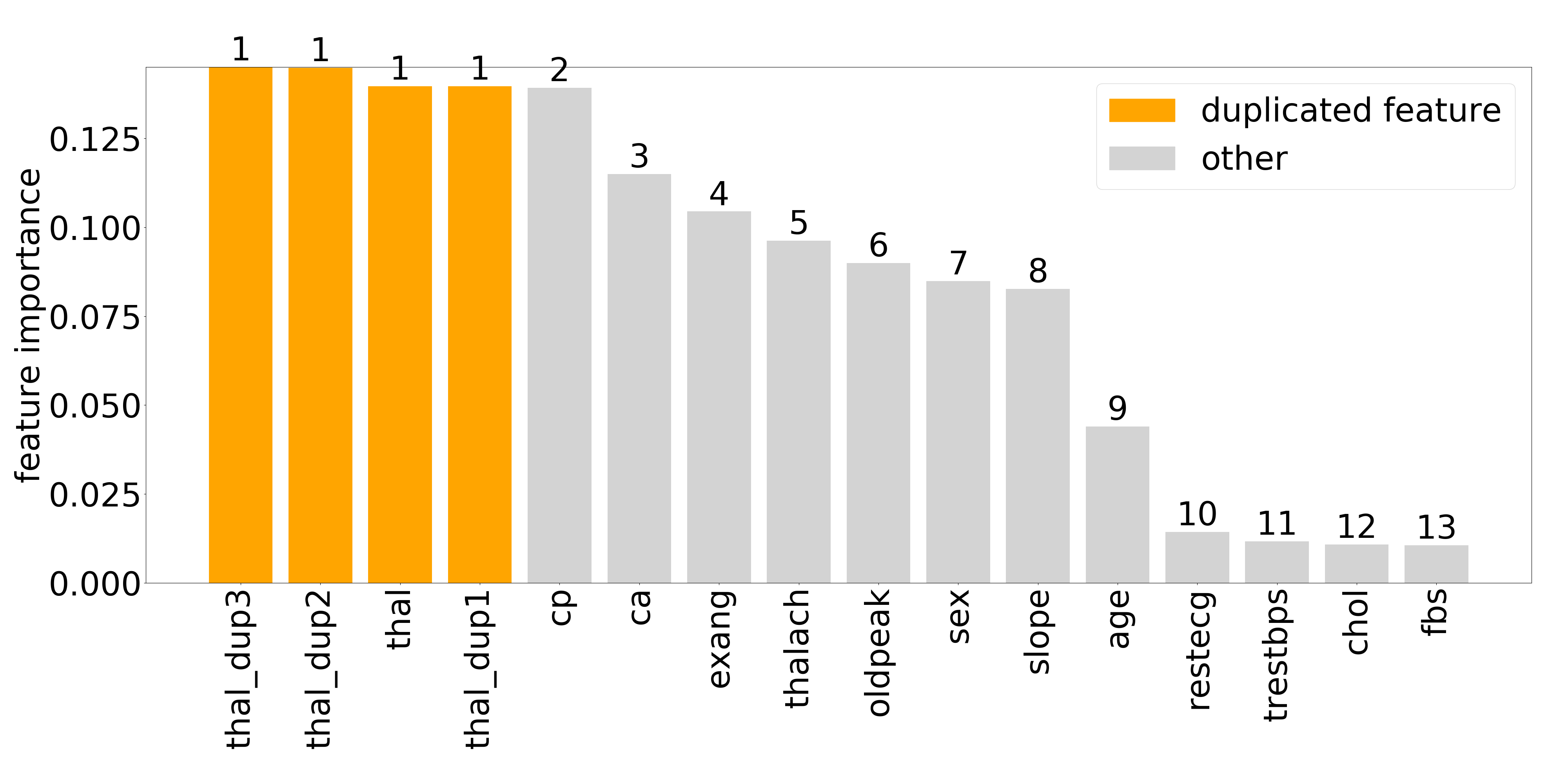}
        \caption{MCI values with duplication}
    \end{subfigure}
    \caption{\textbf{Robustness experiment on the Heart Disease dataset.} On the top row we show the feature importance according to Shapley (a) and MCI (b). Note that that each method suggesting a different ranking within the top three list. The bottom row shows the estimations of both methods, when the top ranked feature of each method is duplicated three times. As you can see, Shapley' importance assignment (c) is affected drastically form the introduction of duplicates, while MCI (d) succeeds to remain stable.}
    \label{fig:heart_dup_experiment}
\end{figure*}

\begin{figure*}[p]
    \centering
    \begin{subfigure}[t]{0.5\textwidth}
        \centering
        \includegraphics[height=1.3in]{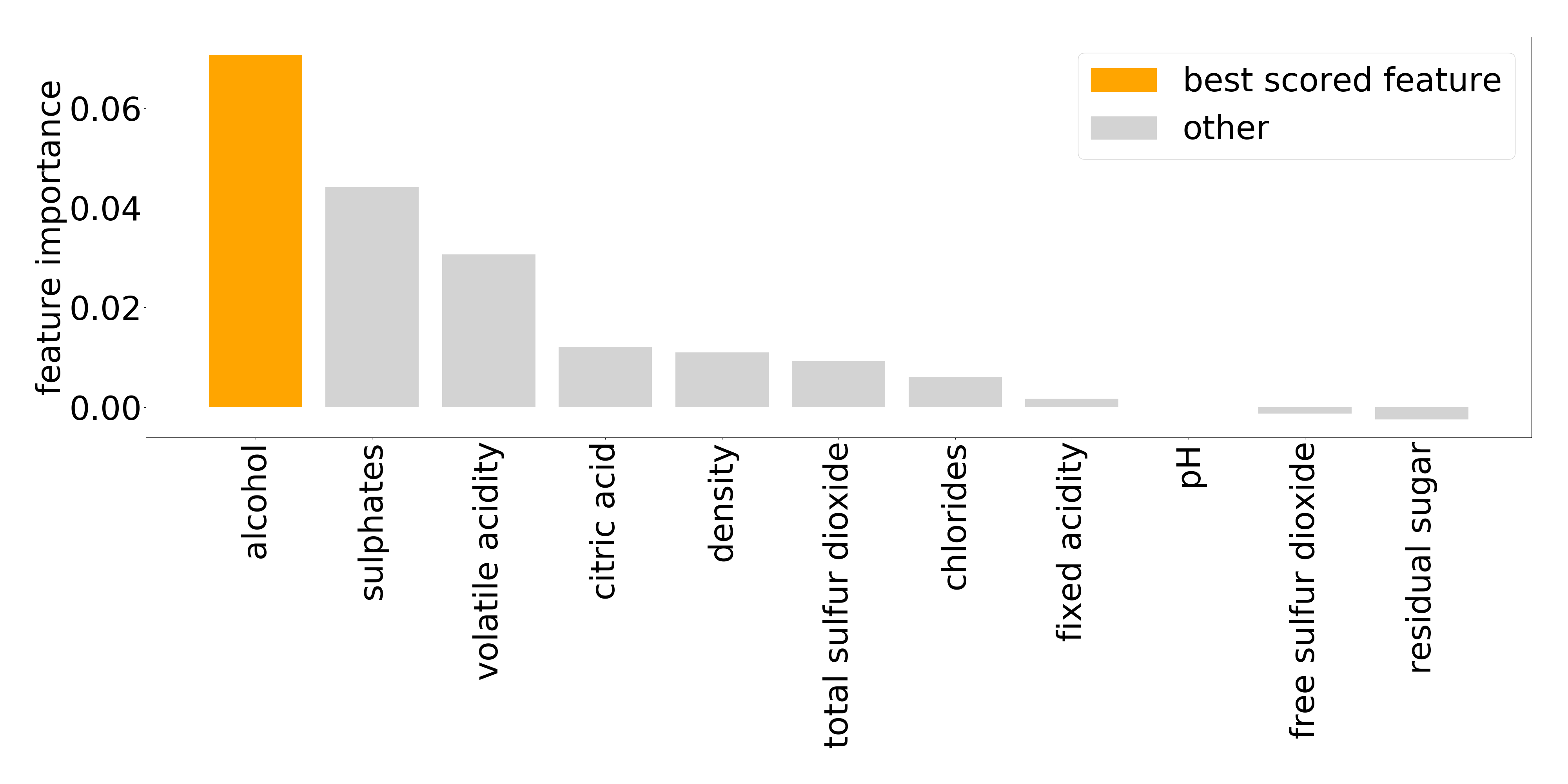}
        \caption{Shapley values}
    \end{subfigure}%
    ~
    \begin{subfigure}[t]{0.5\textwidth}
        \centering
        \includegraphics[height=1.3in]{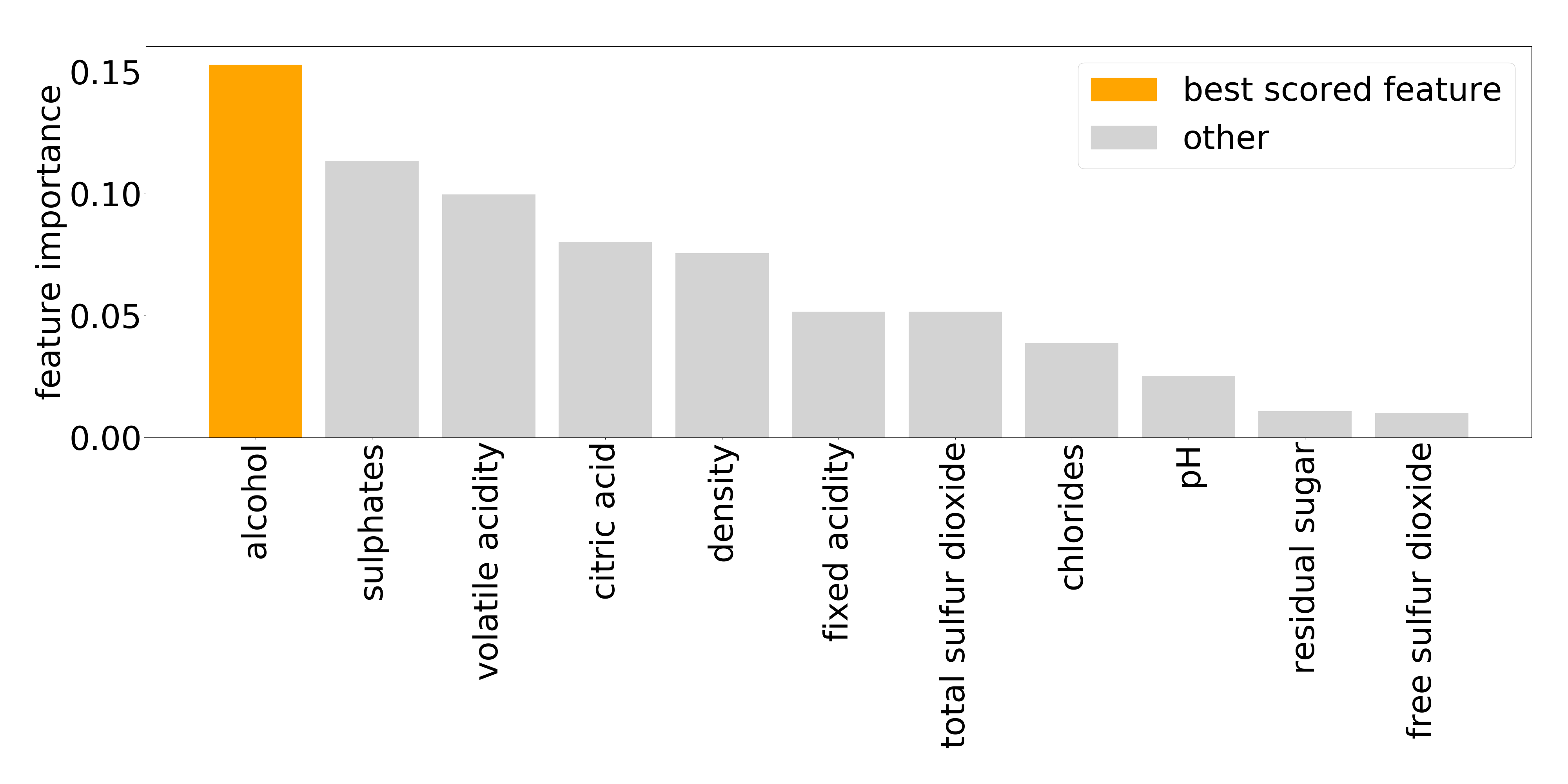}
        \caption{MCI values}
    \end{subfigure}
    \\
    \begin{subfigure}[t]{0.5\textwidth}
        \centering
        \includegraphics[height=1.3in]{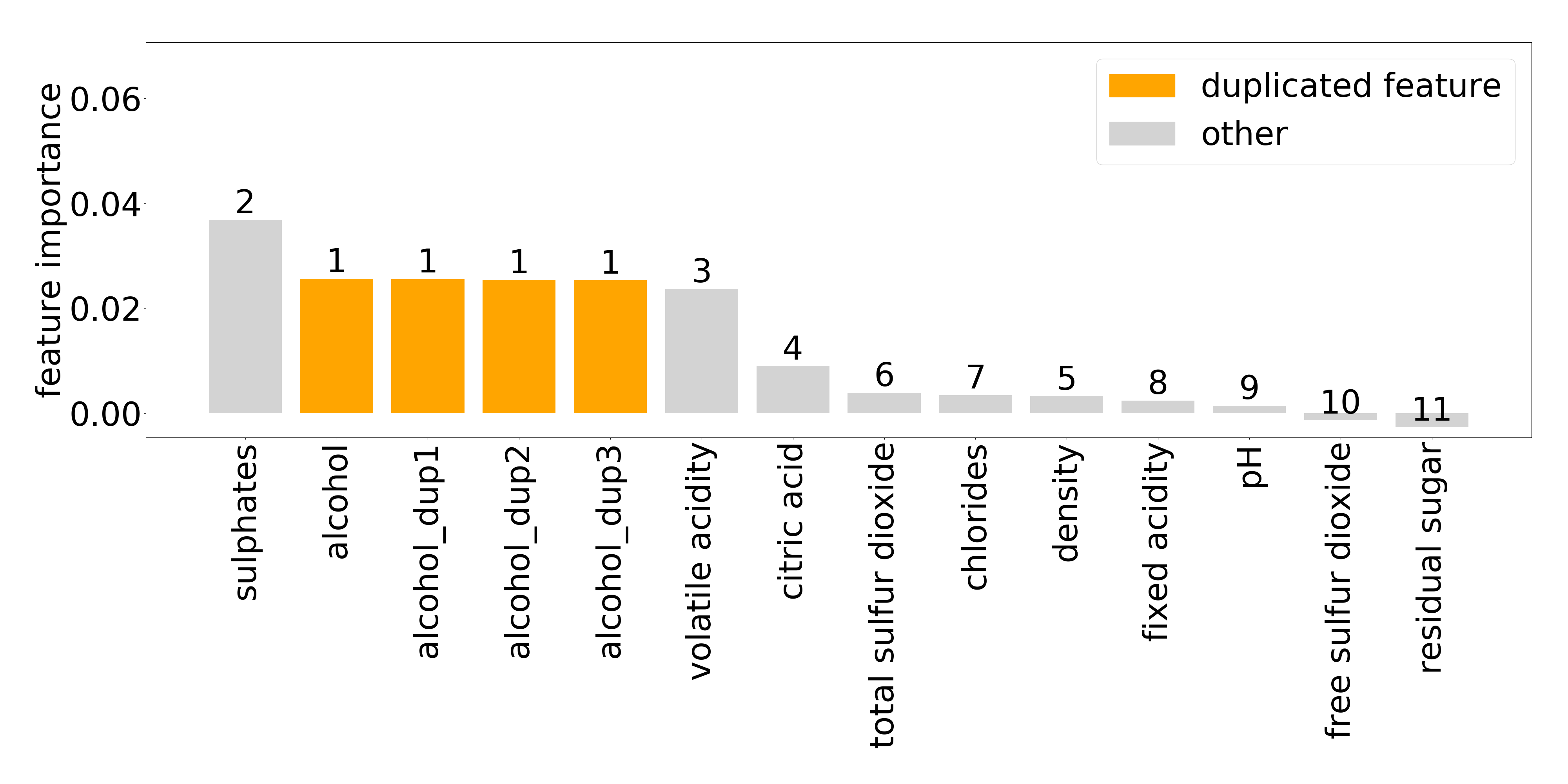}
        \caption{Shapley values with duplication}
    \end{subfigure}%
    ~
    \begin{subfigure}[t]{0.5\textwidth}
        \centering
        \includegraphics[height=1.3in]{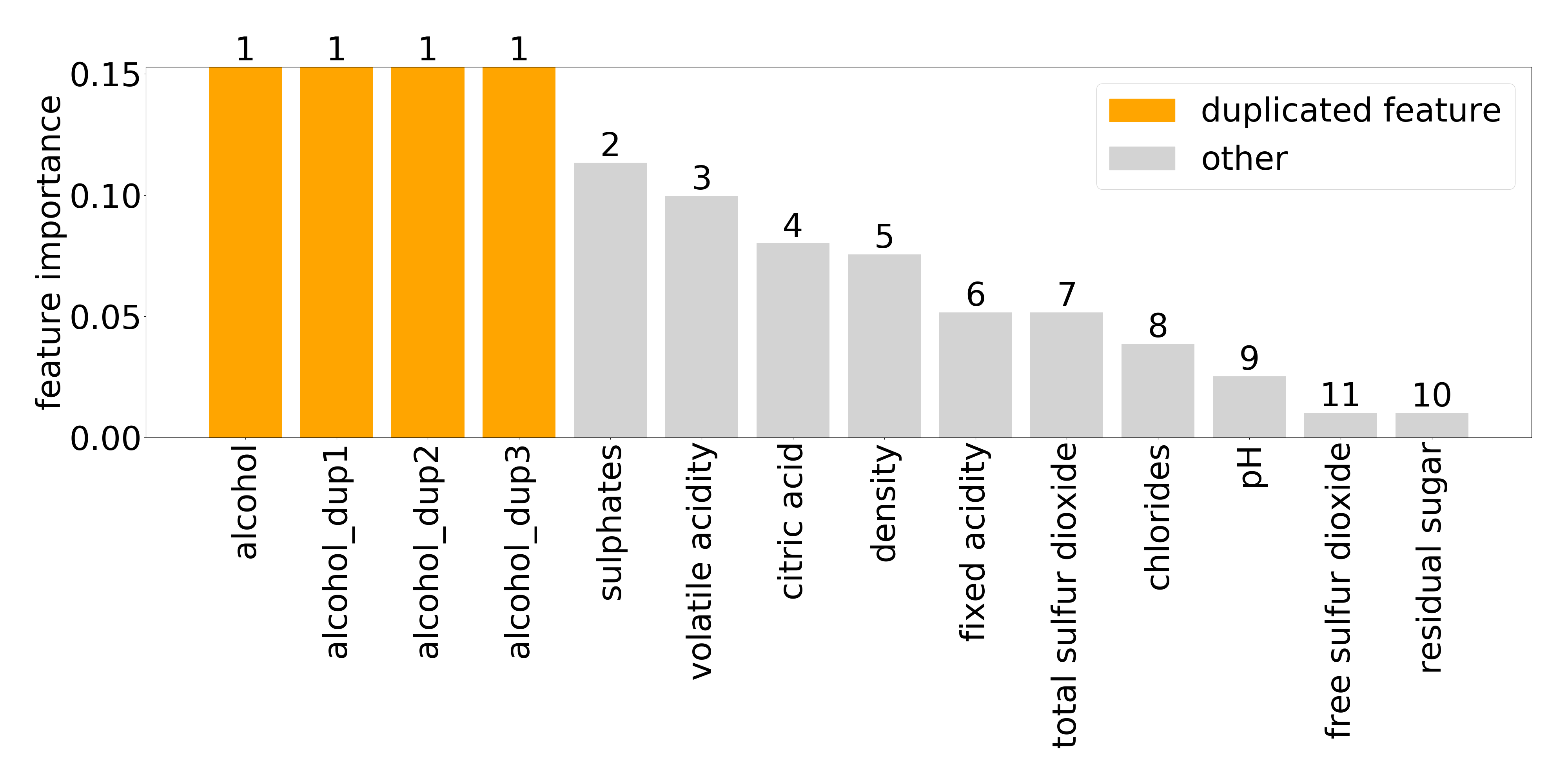}
        \caption{MCI values with duplication}
    \end{subfigure}
    \caption{\textbf{Robustness experiment on the Wine Quality dataset.} On the top row we show the feature importance according to Shapley (a) and MCI (b). The bottom row shows the estimations of both methods, when the top ranked feature of each method is duplicated three times. As you can see, Shapley' importance assignment (c) is affected drastically form the introduction of duplicates, while MCI (d) succeeds to remain stable.}
    \label{fig:wine_dup_experiment}
\end{figure*}

\begin{figure*}[p]
    \centering
    \begin{subfigure}[t]{0.5\textwidth}
        \centering
        \includegraphics[height=1.3in]{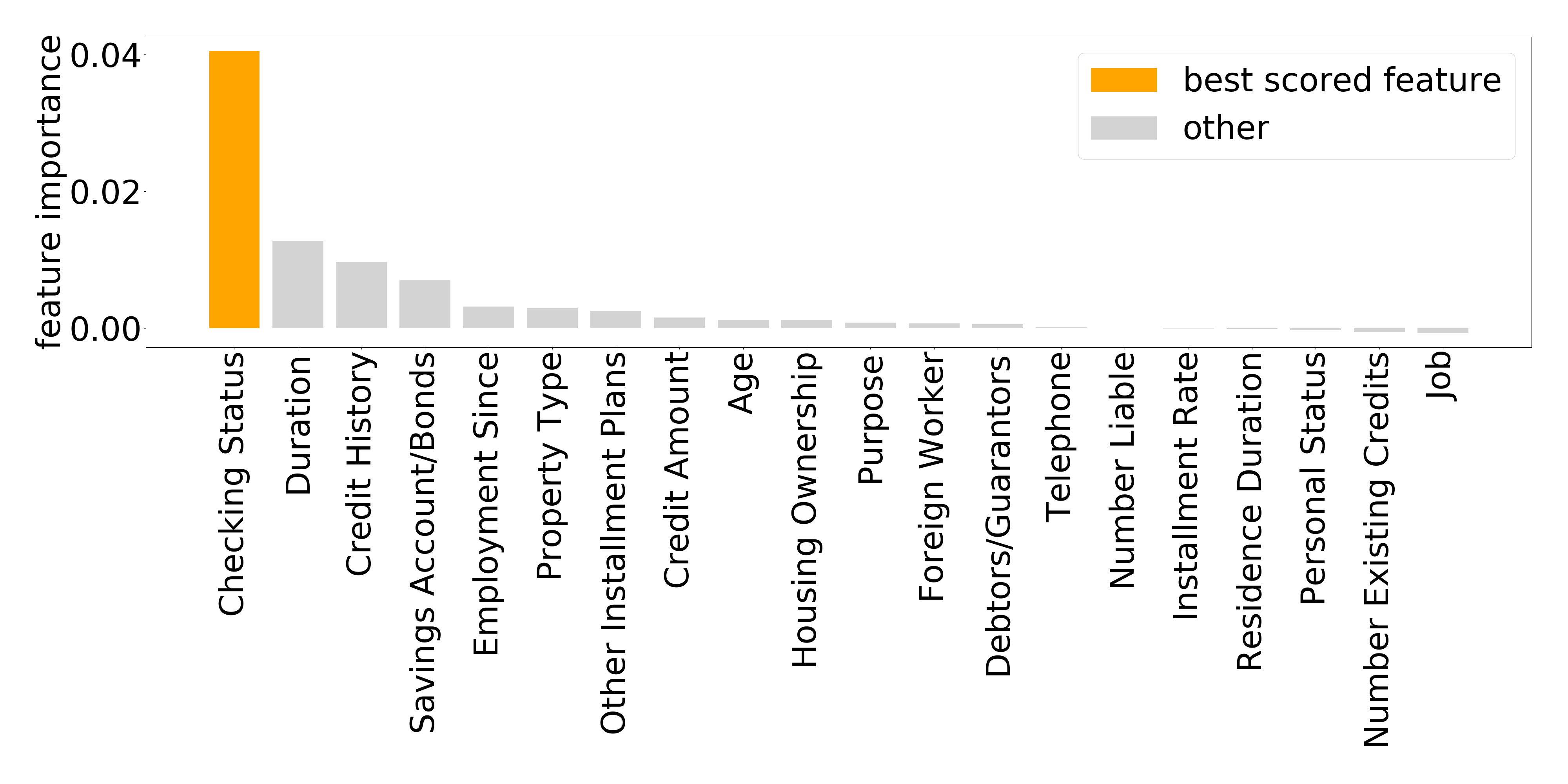}
        \caption{Shapley values}
    \end{subfigure}%
    ~
    \begin{subfigure}[t]{0.5\textwidth}
        \centering
        \includegraphics[height=1.3in]{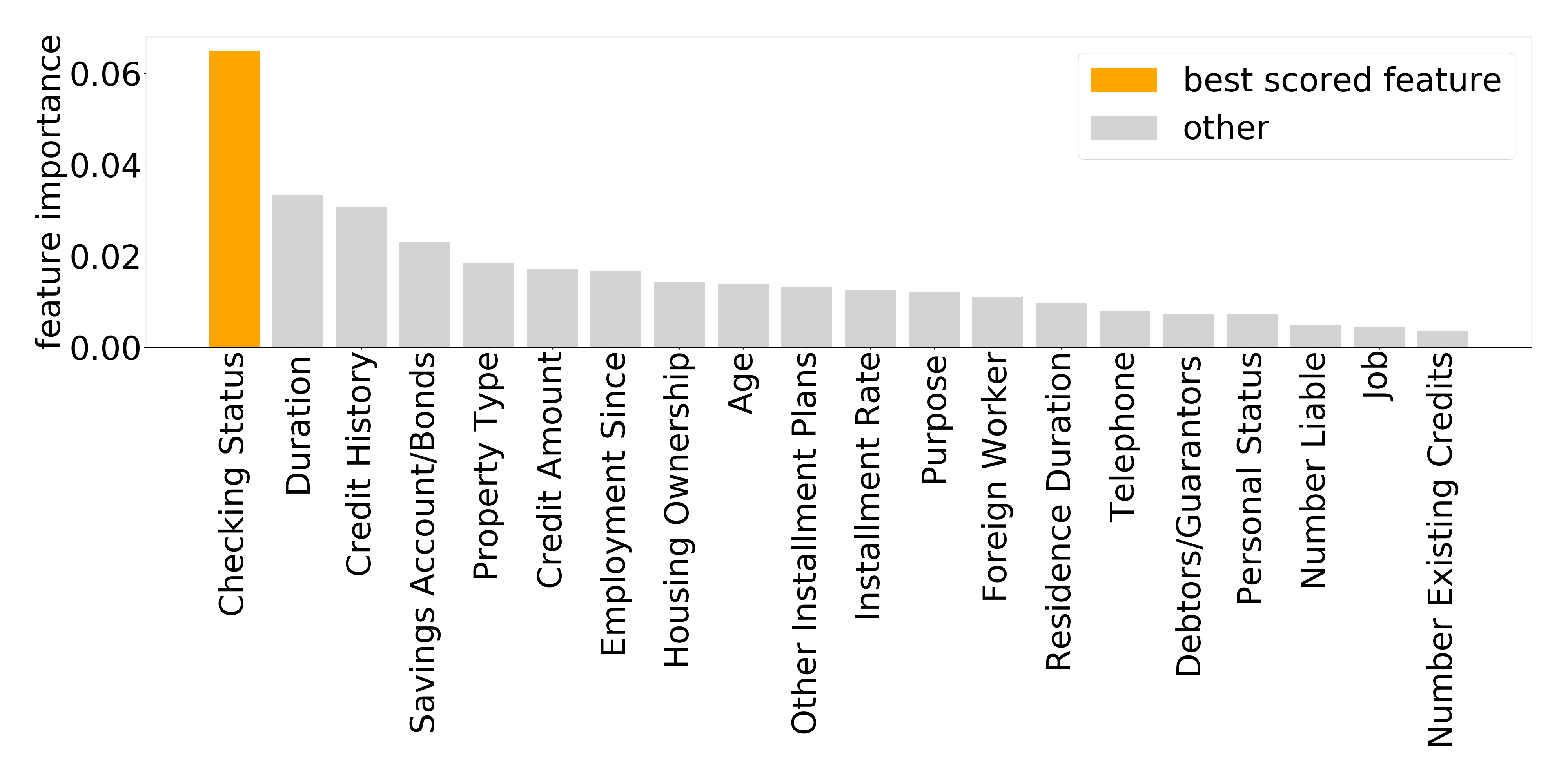}
        \caption{MCI values}
    \end{subfigure}
    \\
    \begin{subfigure}[t]{0.5\textwidth}
        \centering
        \includegraphics[height=1.3in]{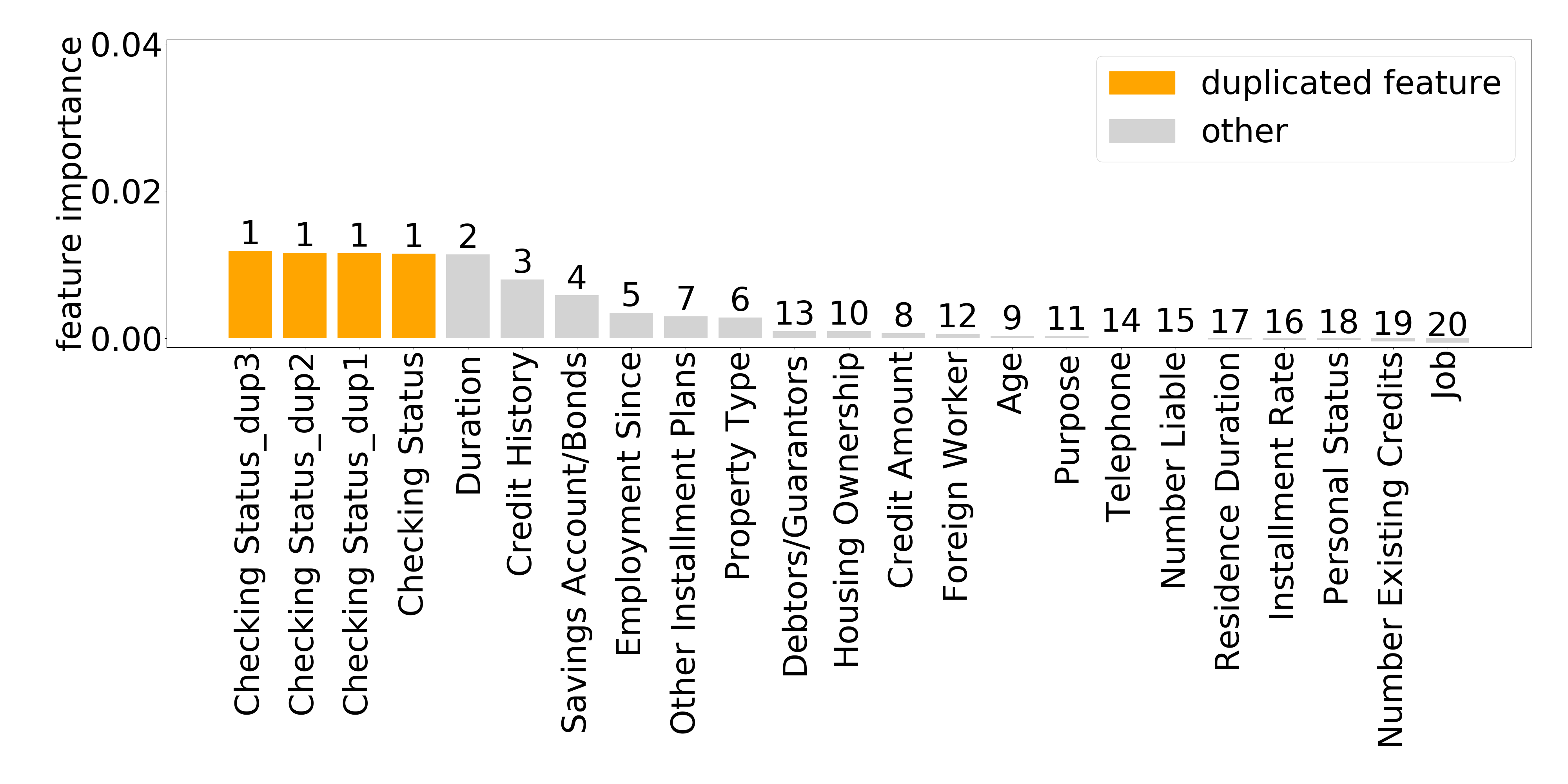}
        \caption{Shapley values with duplication}
    \end{subfigure}%
    ~
    \begin{subfigure}[t]{0.5\textwidth}
        \centering
        \includegraphics[height=1.3in]{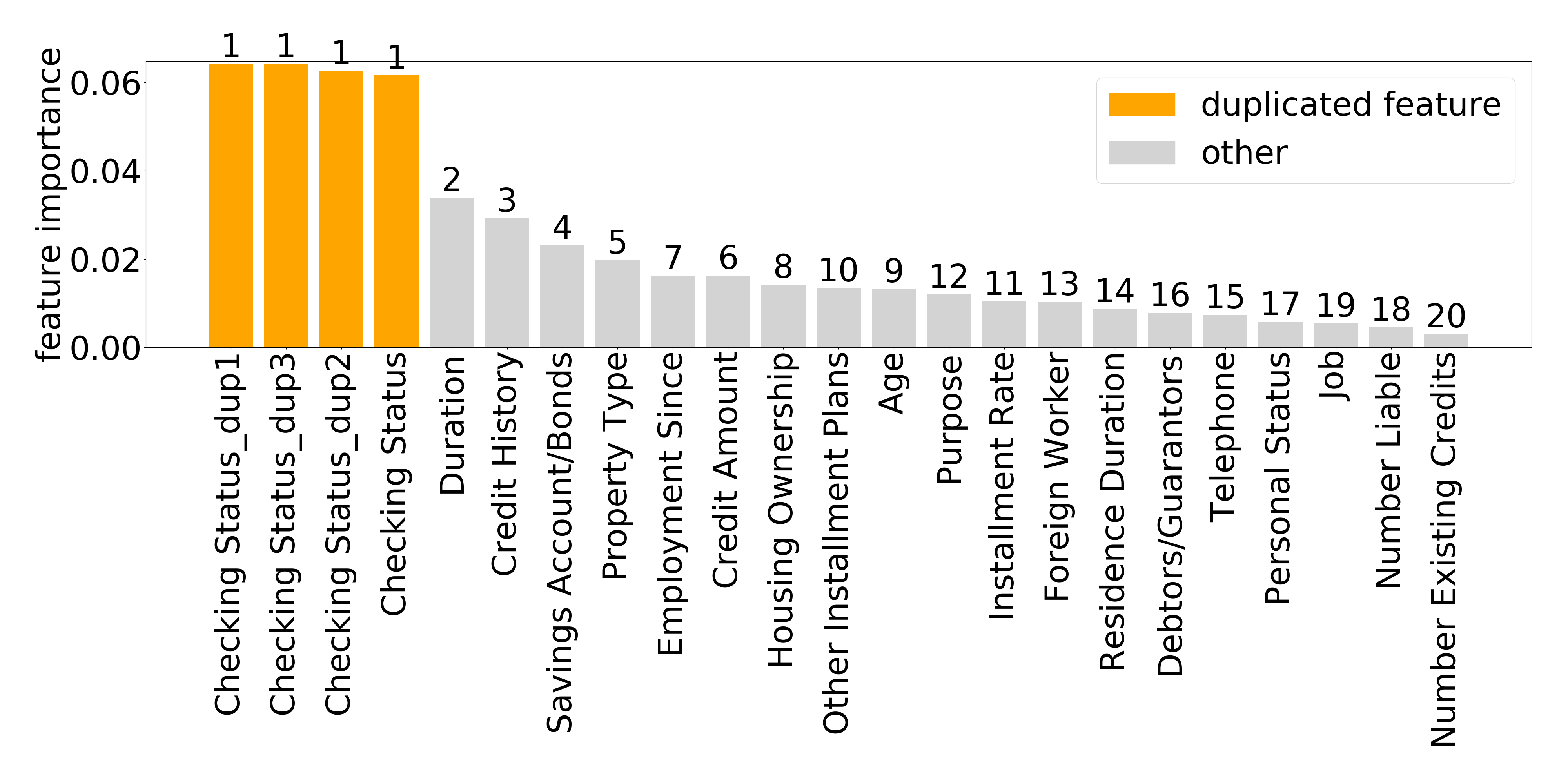}
        \caption{MCI values with duplication}
    \end{subfigure}
    \caption{\textbf{Robustness experiment on the German Credit Default dataset.} On the top row we show the feature importance according to Shapley (a) and MCI (b). The bottom row shows the estimations of both methods, when the top ranked feature of each method is duplicated three times. As you can
    see, the relative differences in the importance scores given by Shapley (c) is affected form the introduction of duplicates, while MCI (d) succeeds to remain stable.}
    \label{fig:german_dup_experiment}
\end{figure*}

\begin{figure*}[p]
    \centering
    \begin{subfigure}[t]{0.5\textwidth}
        \centering
        \includegraphics[height=1.3in]{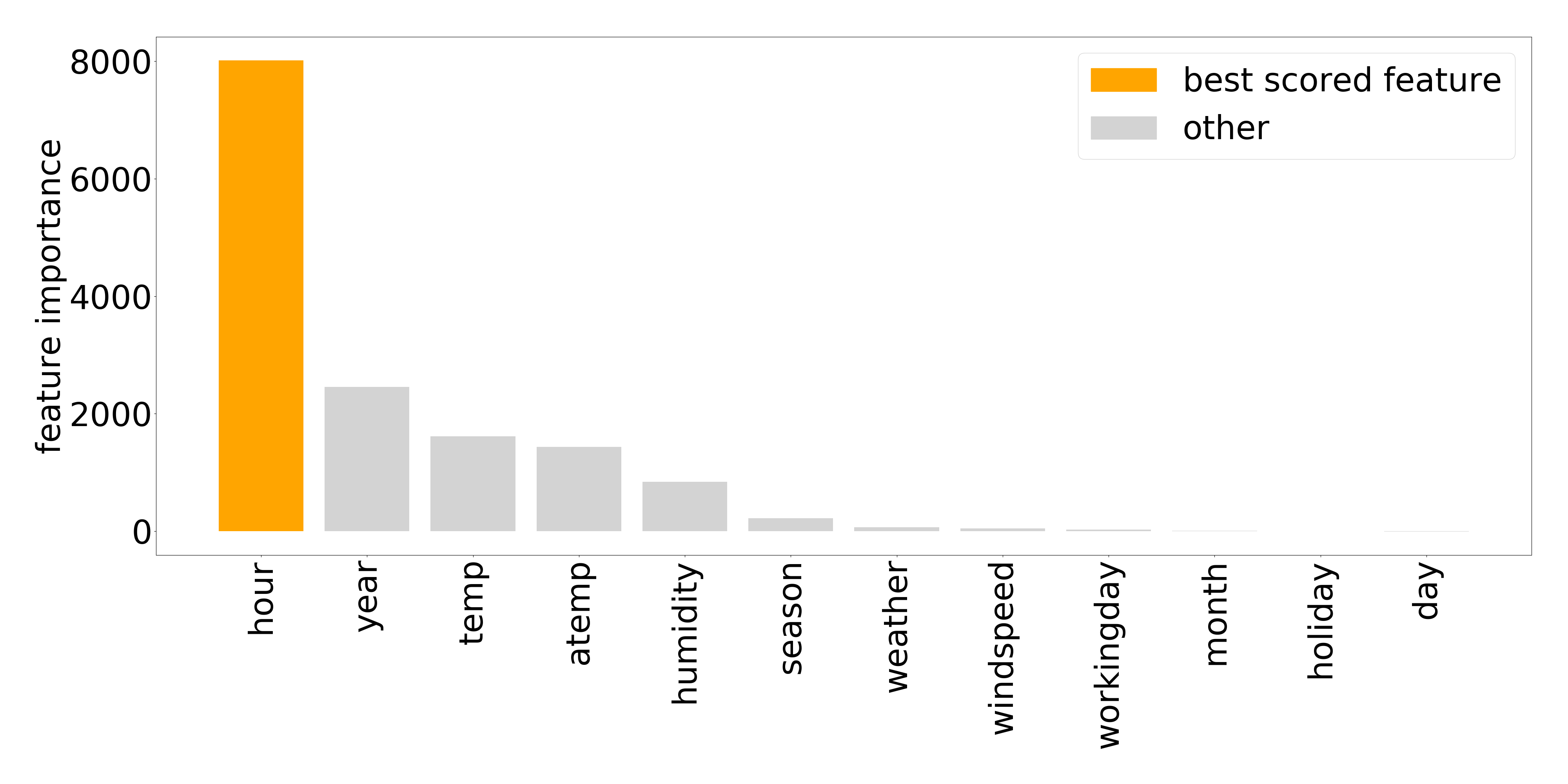}
        \caption{Shapley values}
    \end{subfigure}%
    ~
    \begin{subfigure}[t]{0.5\textwidth}
        \centering
        \includegraphics[height=1.3in]{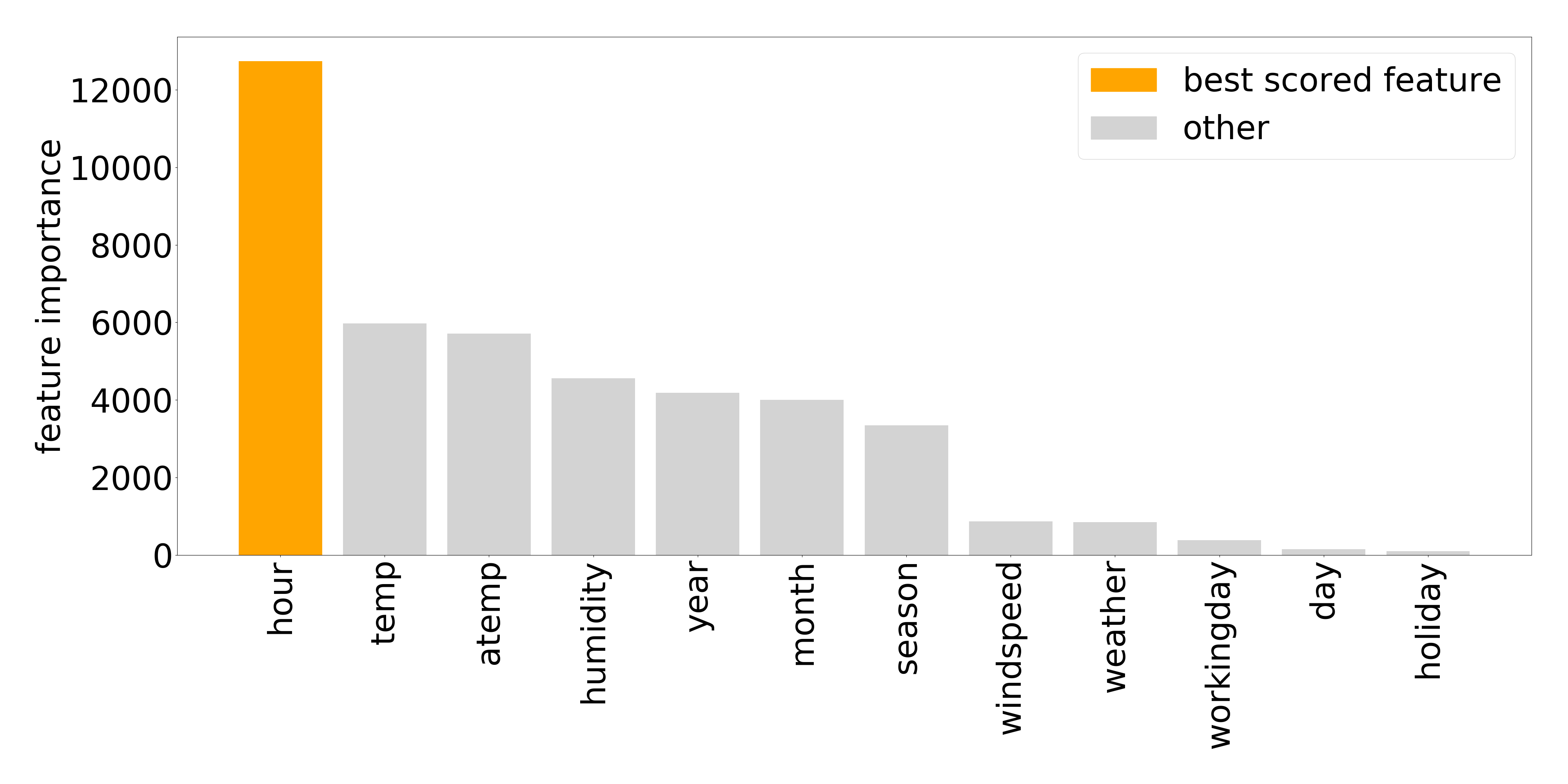}
        \caption{MCI values}
    \end{subfigure}
    \\
    \begin{subfigure}[t]{0.5\textwidth}
        \centering
        \includegraphics[height=1.3in]{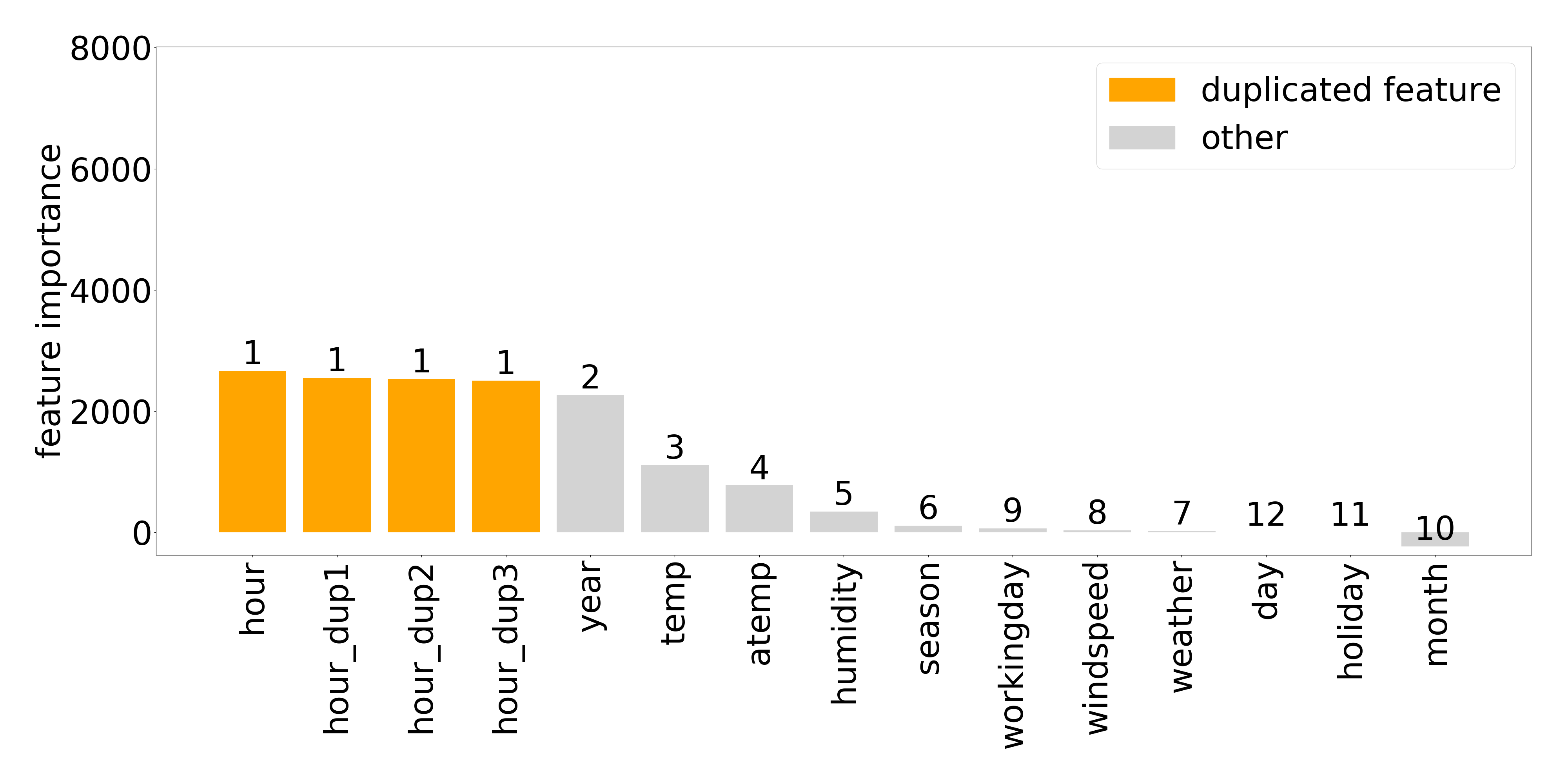}
        \caption{Shapley values with duplication}
    \end{subfigure}%
    ~
    \begin{subfigure}[t]{0.5\textwidth}
        \centering
        \includegraphics[height=1.3in]{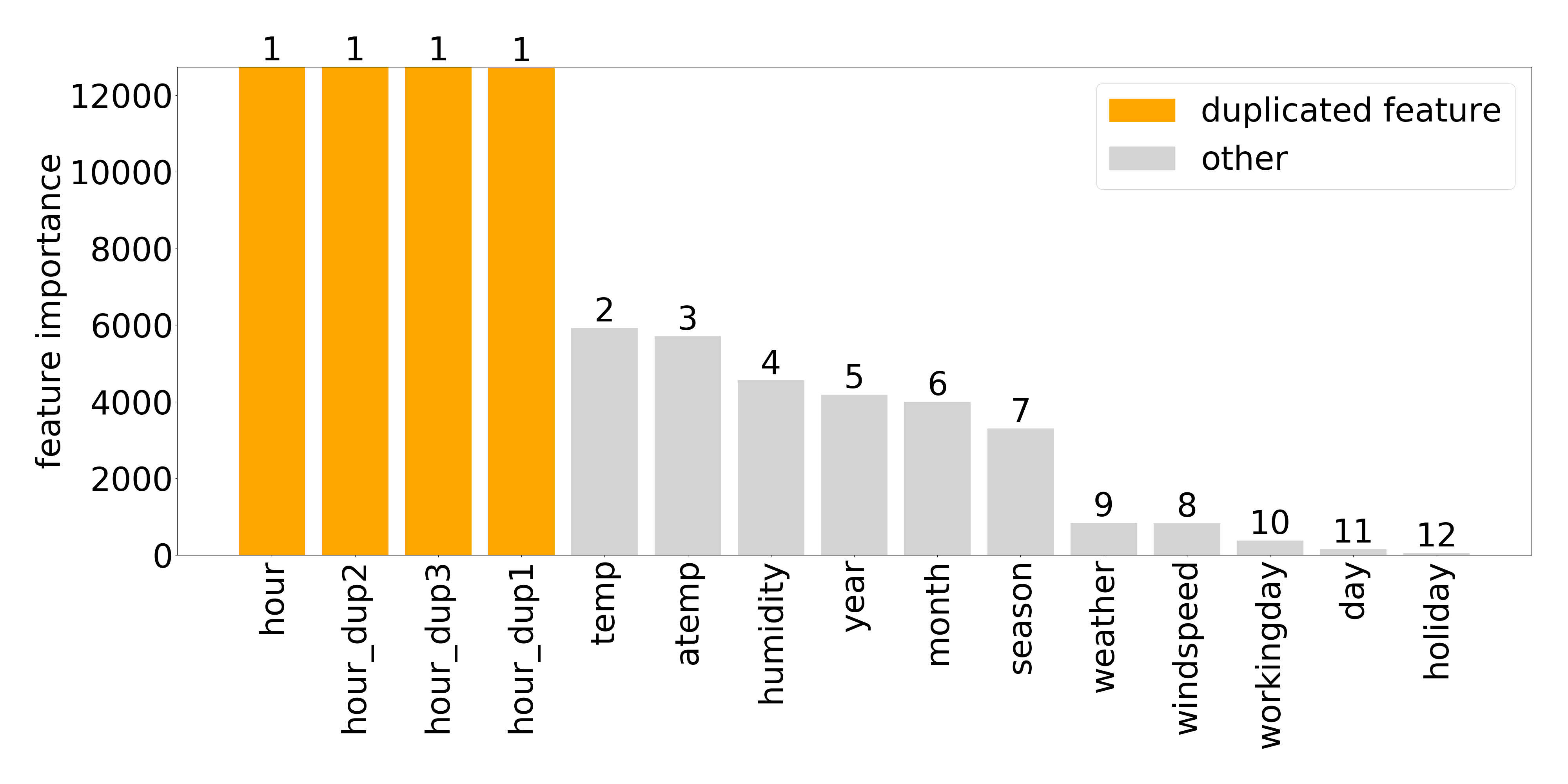}
        \caption{MCI values with duplication}
    \end{subfigure}
    \caption{\textbf{Robustness experiment on the Bike Rental dataset.} On the top row we show the feature importance according to Shapley (a) and MCI (b). The bottom row shows the estimations of both methods, when the top ranked feature of each method is duplicated three times. As you can
    see, the relative differences in the importance scores given by Shapley (c) is affected form the introduction of duplicates, while MCI (d) succeeds to remain stable.}
    \label{fig:bike_dup_experiment}
\end{figure*}

\begin{figure*}[p]
    \centering
    \begin{subfigure}[t]{0.5\textwidth}
        \centering
        \includegraphics[height=1.3in]{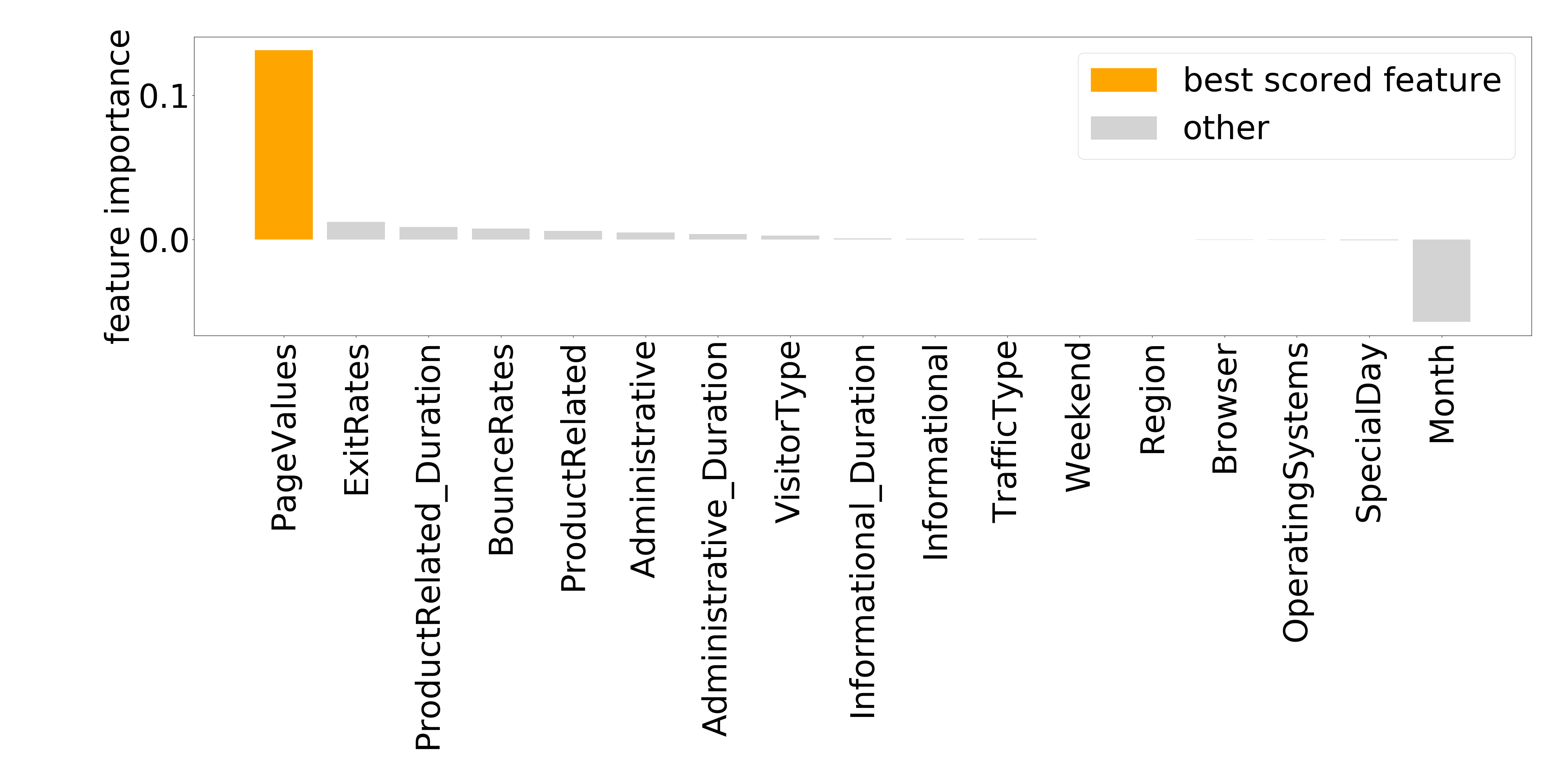}
        \caption{Shapley values}
    \end{subfigure}%
    ~
    \begin{subfigure}[t]{0.5\textwidth}
        \centering
        \includegraphics[height=1.3in]{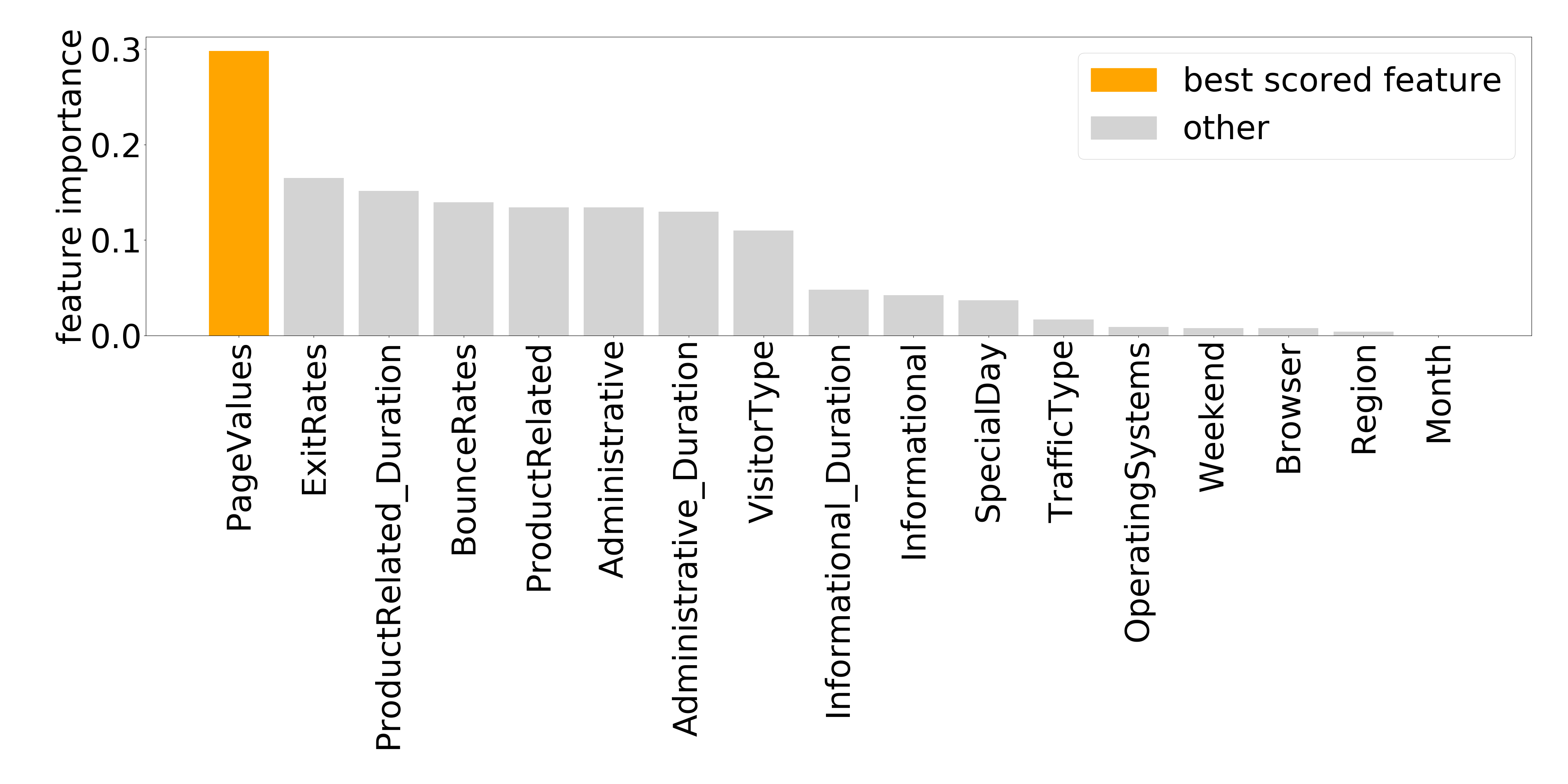}
        \caption{MCI values}
    \end{subfigure}
    \\
    \begin{subfigure}[t]{0.5\textwidth}
        \centering
        \includegraphics[height=1.3in]{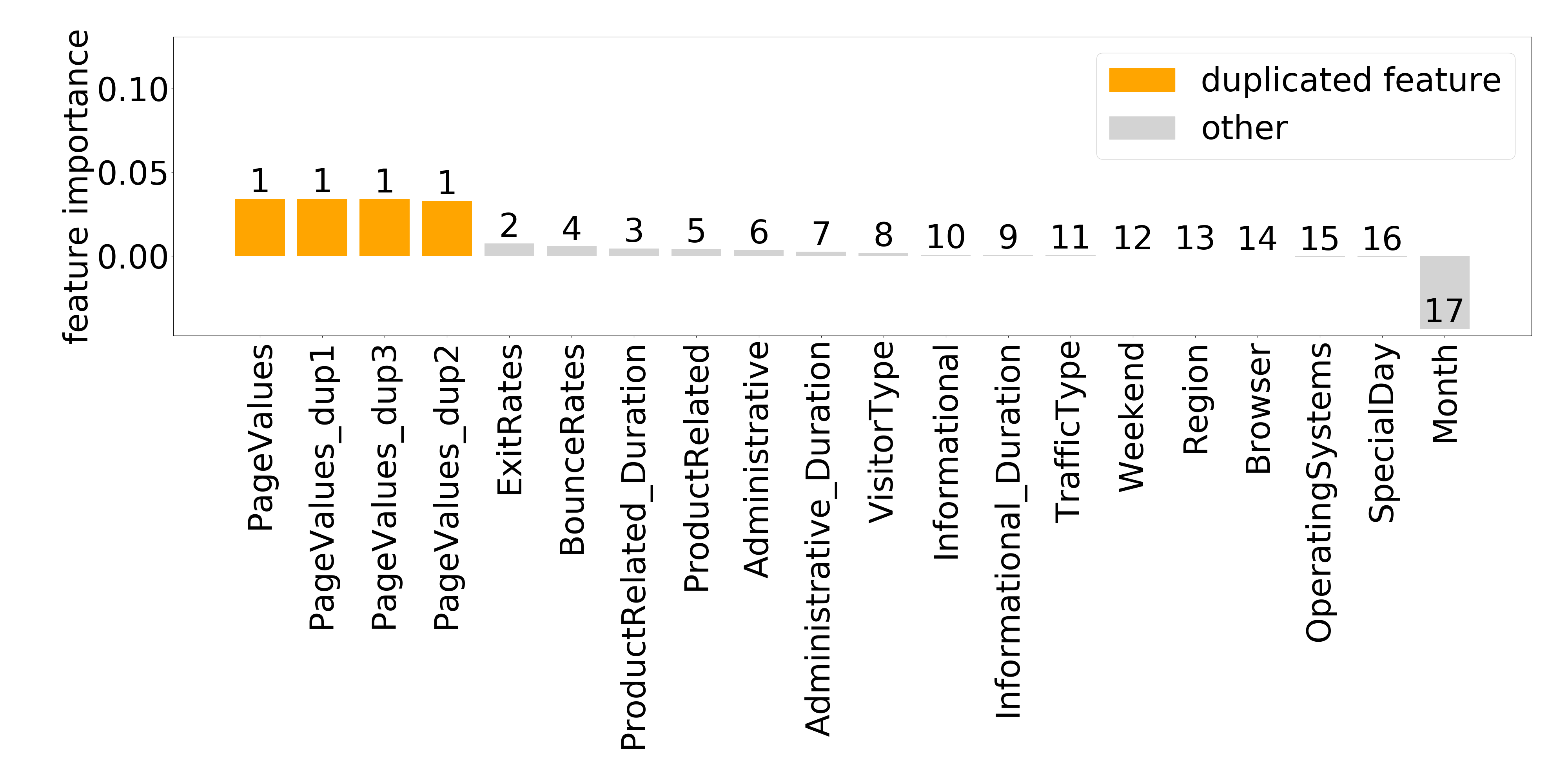}
        \caption{Shapley values with duplication}
    \end{subfigure}%
    ~
    \begin{subfigure}[t]{0.5\textwidth}
        \centering
        \includegraphics[height=1.3in]{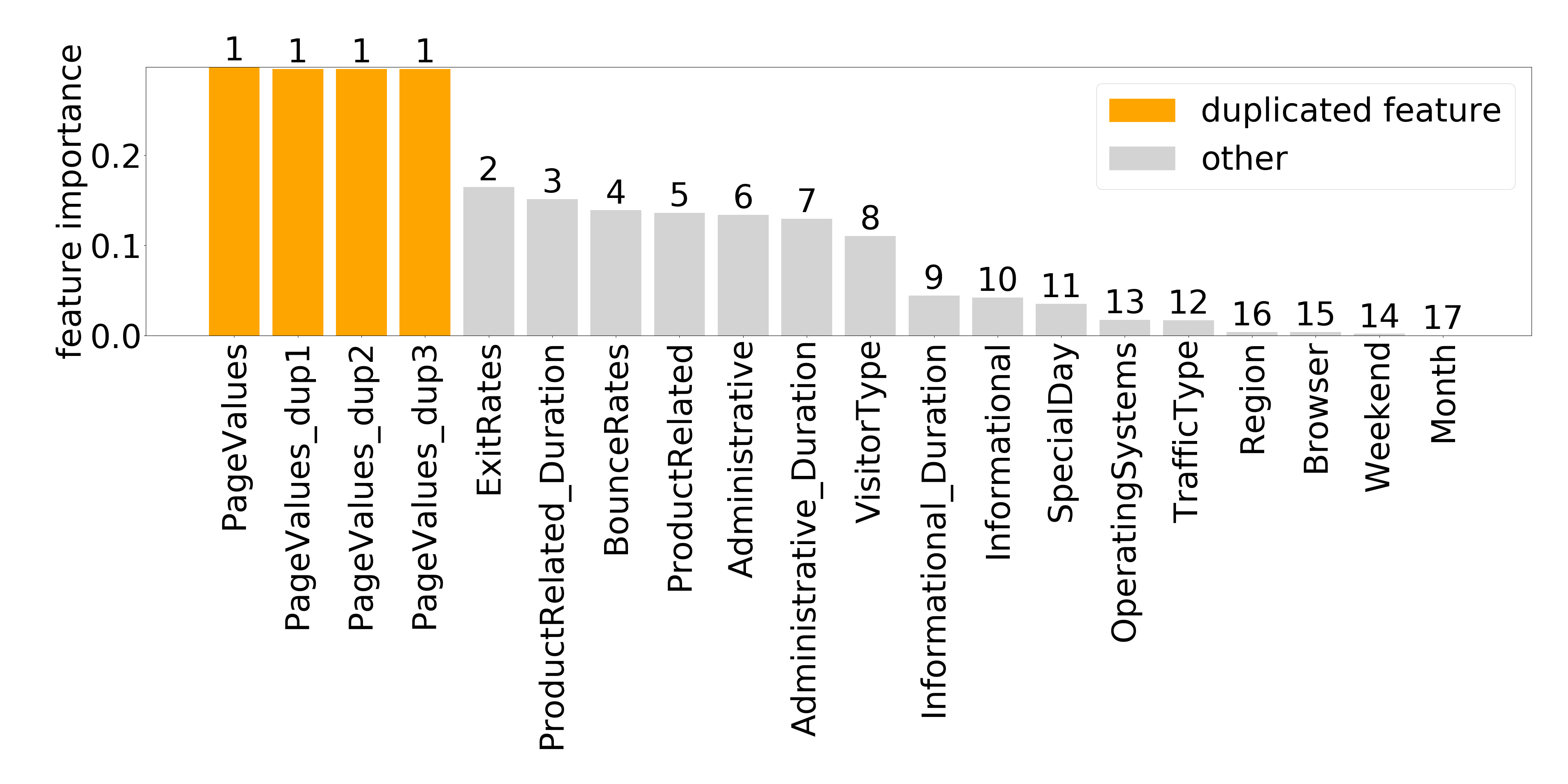}
        \caption{MCI values with duplication}
    \end{subfigure}
    \caption{\textbf{Robustness experiment on the Online Shopping dataset.} On the top row we show the feature importance according to  Shapley (a) and MCI (b). The bottom row shows the estimations of both methods, when the top ranked feature of each method is duplicated three times. As you can
    see, the relative differences in the importance scores given by Shapley (c) is affected form the introduction of duplicates, while MCI (d) succeeds to remain stable.}
    \label{fig:shopping_dup_experiment}
\end{figure*}

\begin{figure*}[p]
    \centering
    \begin{subfigure}[t]{0.5\textwidth}
        \centering
        \includegraphics[height=1.3in]{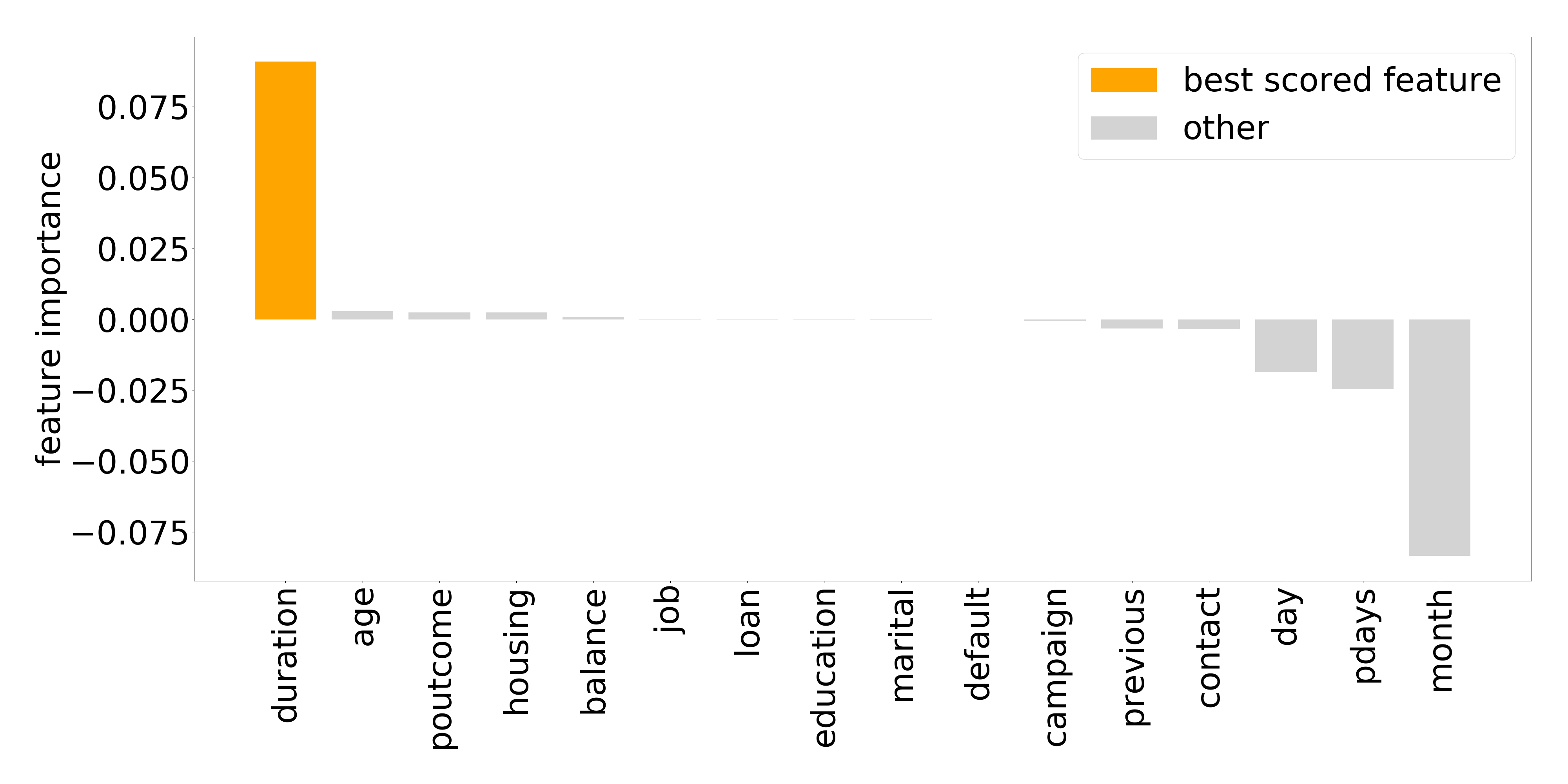}
        \caption{Shapley values}
    \end{subfigure}%
    ~
    \begin{subfigure}[t]{0.5\textwidth}
        \centering
        \includegraphics[height=1.3in]{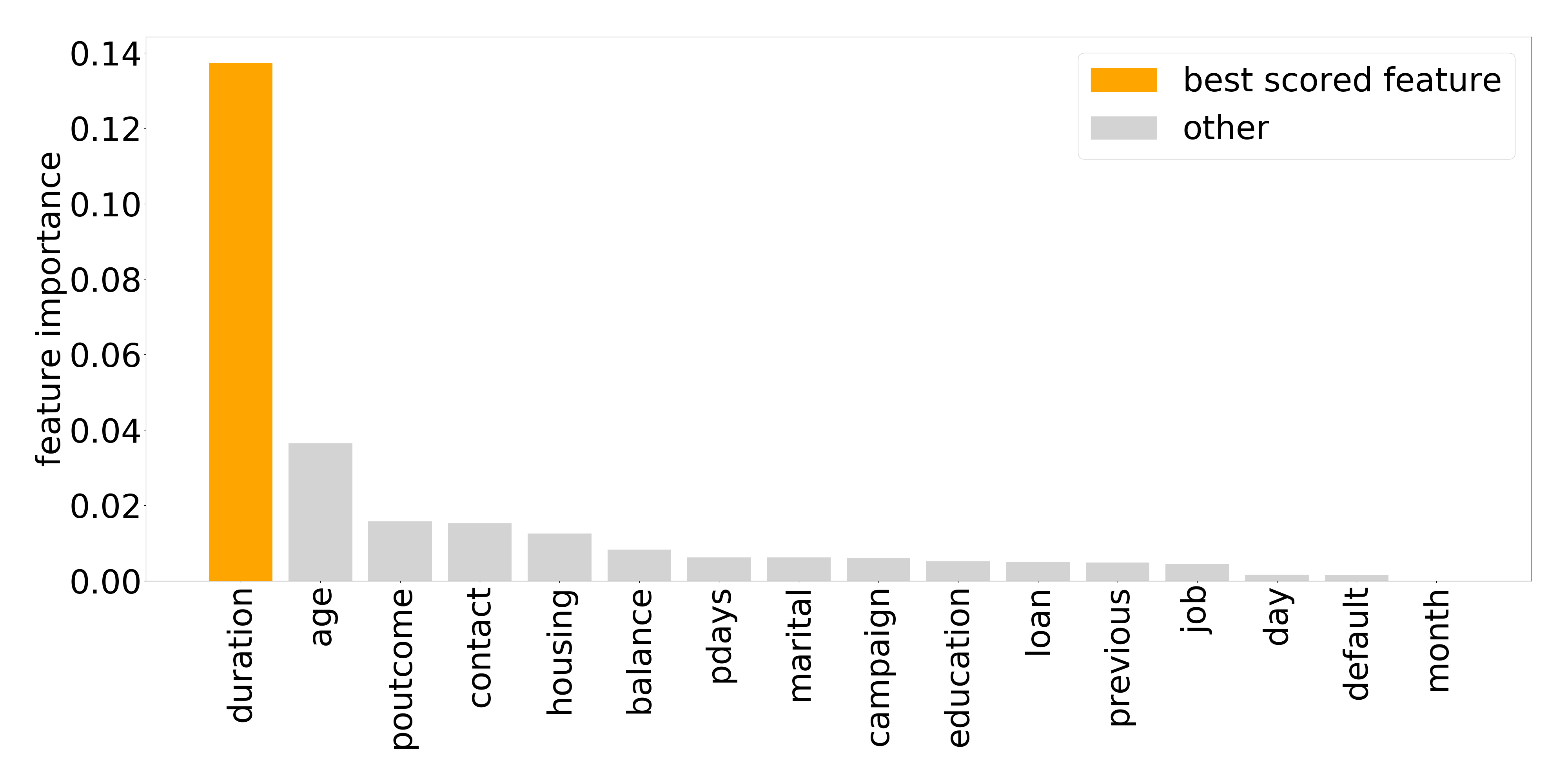}
        \caption{MCI values}
    \end{subfigure}
    \\
    \begin{subfigure}[t]{0.5\textwidth}
        \centering
        \includegraphics[height=1.3in]{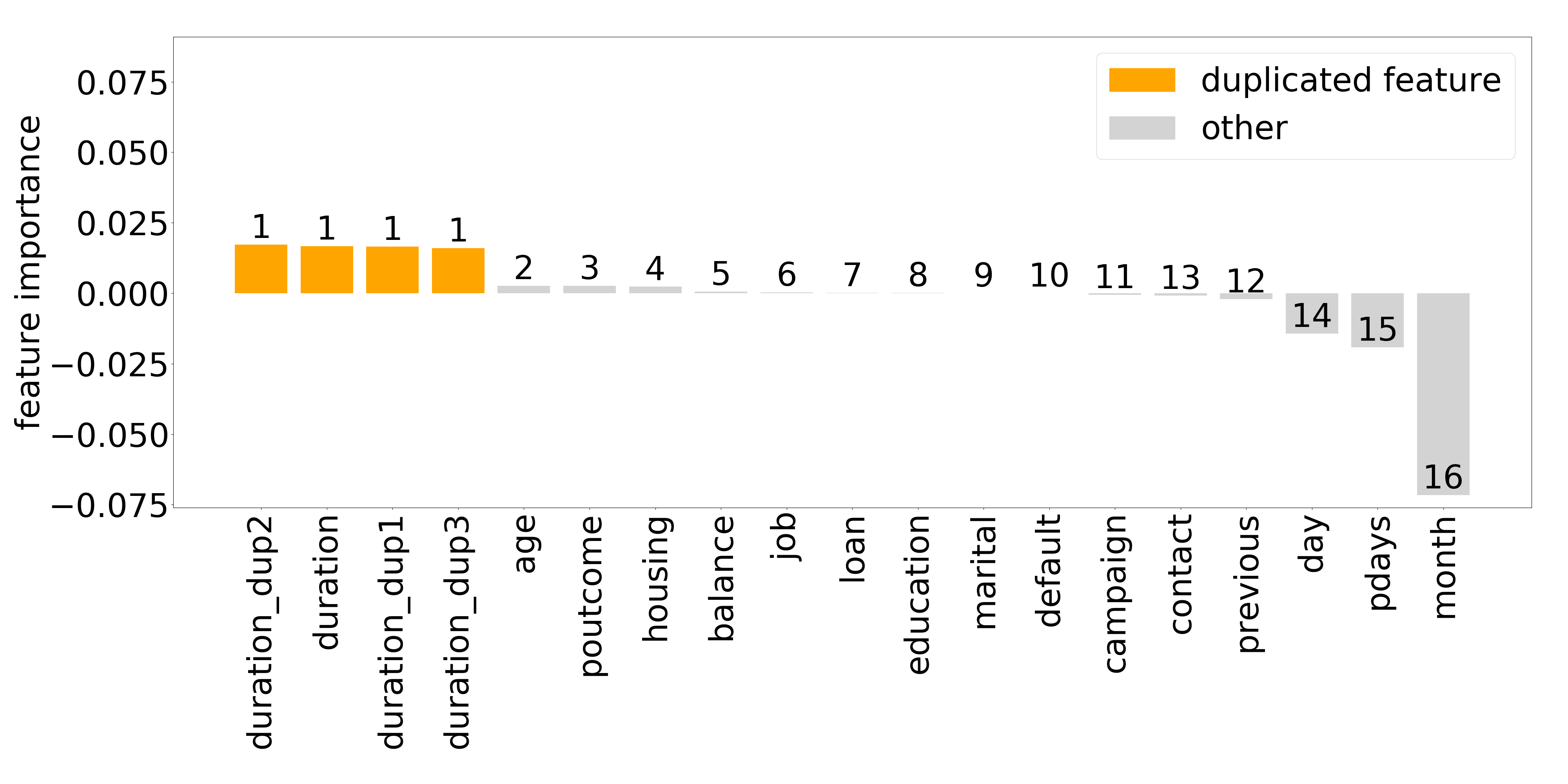}
        \caption{Shapley values with duplication}
    \end{subfigure}%
    ~
    \begin{subfigure}[t]{0.5\textwidth}
        \centering
        \includegraphics[height=1.3in]{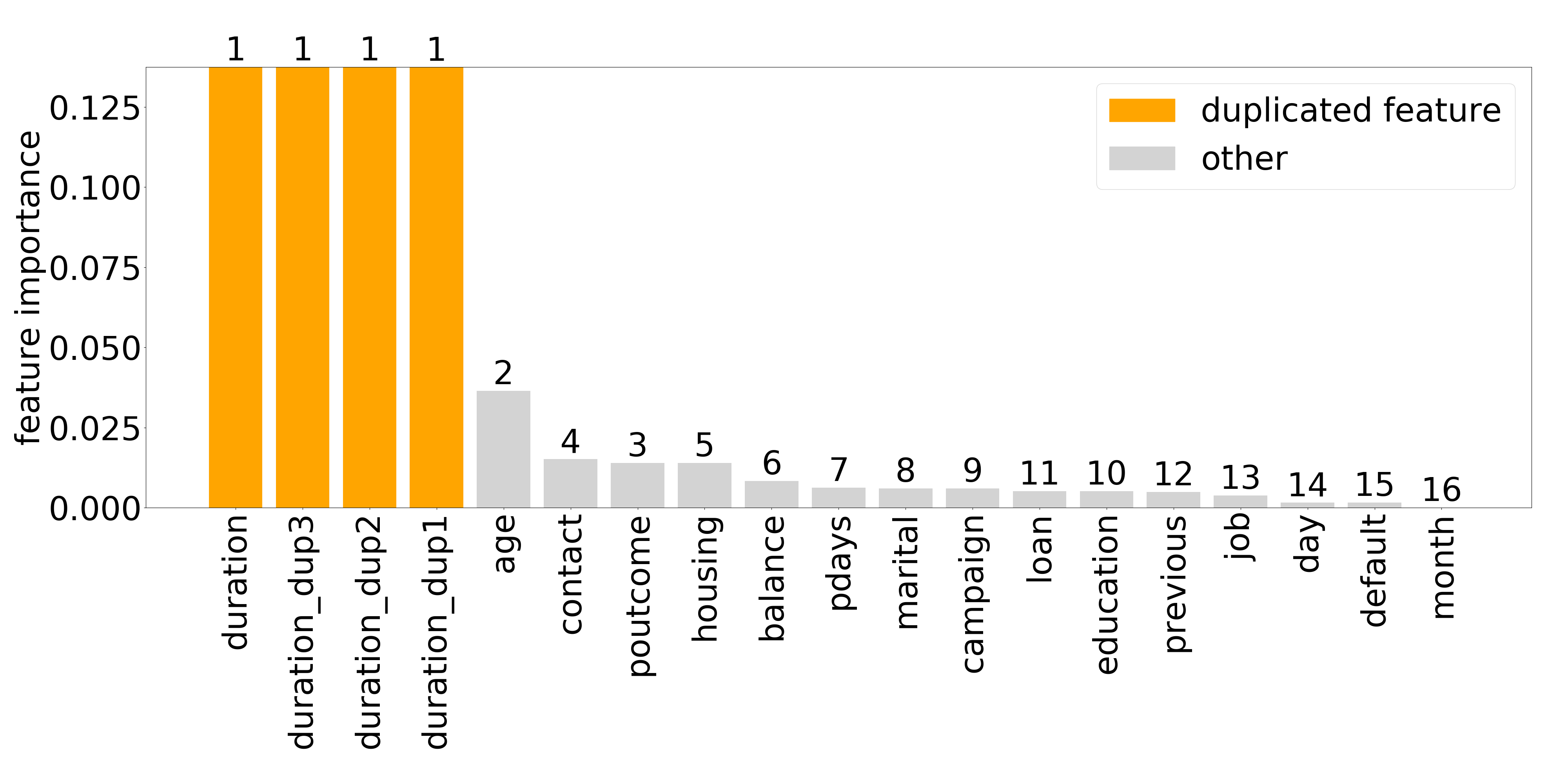}
        \caption{MCI values with duplication}
    \end{subfigure}
    \caption{\textbf{Robustness experiment on the Bank Marketing dataset.} On the top row we show the feature importance according  Shapley (a) and MCI (b). The bottom row shows the estimations of both methods, when the top ranked feature of each method is duplicated three times. As you can
    see, the relative differences in the importance scores given by Shapley (c) is affected form the introduction of duplicates, while MCI (d) succeeds to remain stable.}
    \label{fig:bank_dup_experiment}
\end{figure*}

\begin{table}[p]
\centering
\caption{{\bf{Results of the robustness experiments for the Heart Disease dataset}:  \textmd{Below are the Minimal Kendall Distance (MKD) between the rankings of the top-$k$ ranked elements  before and after duplicating the top feature three times. In these results lower is better, the perfect score is 0.00. As shown here, Shapley values was highly affected from the introduction to duplicates, while MCI remained stable.}}}.
\begin{tabular}{lcccc}
\toprule{} Method   &     @3 &       @5 &      @10 &          @13 \\
\midrule        MCI &  \textbf{0.00} &  \textbf{0.00} &  \textbf{0.00} &  \textbf{0.00}\\

    Shapley &  0.67 &  0.18 &  0.13 &  0.07 \\
    Ablation &  0.13 &  0.11 &  0.11 &  0.06 \\
    Bivariate &  \textbf{0.00} &  \textbf{0.00} &  \textbf{0.00} &  \textbf{0.00} \\
\bottomrule
\label{table:UCI_heart_kendall}
\end{tabular}
\end{table}
\begin{table}[p]
\centering
\caption{{\bf{Results of the robustness experiments for the Wine Quality dataset}:  \textmd{Below are the Minimal Kendall Distance (MKD) between the rankings of the top-$k$ ranked elements  before and after duplicating the top feature three times. In these results lower is better, the perfect score is 0.00. As shown here, Shapley values was highly affected from the introduction to duplicates, while MCI remained stable.}}}.
\begin{tabular}{lcccc}
\toprule{} Method   &     @3 &       @5 &        @11 \\
\midrule
MCI &  \textbf{0.00} &  \textbf{0.00} &  0.18\\
Shapley &  0.33 &  0.07 &  0.05  \\
Ablation &  0.13 &  0.11 &  0.2  \\
Bivariate &  \textbf{0.00} &  \textbf{0.00} &  \textbf{0.00} \\
\bottomrule
\label{table:UCI_wine_kendall}
\end{tabular}
\end{table}
\begin{table}[p]
\centering
\caption{{\bf{Results of the robustness experiments for the  The German Credit Default dataset}:  \textmd{Below are the Minimal Kendall Distance (MKD) between the rankings of the top-$k$ ranked elements  before and after duplicating the top feature three times. In these results lower is better, the perfect score is 0.00. In figure~\ref{fig:german_dup_experiment} we show that the relative score of the top feature was highly affected when using Shapley value, but remains stable when using MCI.}}}.
\begin{tabular}{lcccc}
\toprule{} Method   &     @3 &       @5 &      @10 &          @20 \\
\midrule        MCI &  \textbf{0.00} &  \textbf{0.00} &  0.04 &  0.02\\

    Shapley & \textbf{0.00} &  \textbf{0.00} &  0.06 &  0.06\\
    Ablation &  0.13 &  0.11 &  0.07 &  0.08 \\
    Bivariate &  \textbf{0.00} &  \textbf{0.00} &  \textbf{0.00} &  \textbf{0.00} \\
\bottomrule
\label{table:UCI_german_kendall}
\end{tabular}
\end{table}
\begin{table}[p]
\centering
\caption{{\bf{Results of the robustness experiments for the Bike Rental dataset}:  \textmd{Below are the Minimal Kendall Distance (MKD) between the rankings of the top-$k$ ranked elements  before and after duplicating the top feature three times. In these results lower is better, the perfect score is 0.00. In figure~\ref{fig:bike_dup_experiment} we show that the relative score of the top feature was highly affected when using Shapley value, but remains stable when using MCI.}}}.
\begin{tabular}{lcccc}
\toprule{} Method   &     @3 &       @5 &      @10 &          @12 \\
\midrule        MCI &  \textbf{0.00} &  \textbf{0.00} &  0.02 &  0.02\\

    Shapley &  \textbf{0.00} &  \textbf{0.00} &  0.05 &  0.09 \\
    Ablation &  0.13 &  0.11 &  0.22 &  0.15 \\
    Bivariate &  \textbf{0.00} &  \textbf{0.00} &  \textbf{0.00} &  \textbf{0.00} \\
\bottomrule
\label{table:UCI_bike_kendall}
\end{tabular}
\end{table}
\begin{table}[p]
\centering
\caption{{\bf{Results of the robustness experiments for the Online Shopping dataset}:  \textmd{Below are the Minimal Kendall Distance (MKD) between the rankings of the top-$k$ ranked elements  before and after duplicating the top feature three times. In these results lower is better, the perfect score is 0.00. In figure~\ref{fig:shopping_dup_experiment} we show that the relative score of the top feature was affected when using Shapley value, but remains stable when using MCI.}}}.
\begin{tabular}{lcccc}
\toprule{} Method   &     @3 &       @5 &      @10 &          @17 \\
\midrule        MCI &  \textbf{0.00} &  0.10 &  0.02 &  0.04\\

    Shapley &  \textbf{0.00} &  0.06 &  0.07 &  0.02 \\
    Ablation &  0.13 &  0.11 &  0.07 &  0.10 \\
    Bivariate &  \textbf{0.00} &  \textbf{0.00} &  \textbf{0.00} &  \textbf{0.00} \\
\bottomrule
\label{table:UCI_shopping_kendall}
\end{tabular}
\end{table}
\begin{table}[p]
\centering
\caption{{\bf{Results of the robustness experiments for the Bank Marketing dataset}:  \textmd{Below are the Minimal Kendall Distance (MKD) between the rankings of the top-$k$ ranked elements  before and after duplicating the top feature three times. In these results lower is better, the perfect score is 0.00. In figure~\ref{fig:bank_dup_experiment} we show that the relative score of the top feature was affected when using Shapley value, but remains stable when using MCI.}}}.
\begin{tabular}{lcccc}
\toprule{} Method   &     @3 &       @5 &      @10 &          @17 \\
\midrule        MCI &  \textbf{0.00} &  0.10 &  0.02 &  0.02\\

    Shapley &  \textbf{0.00} &  \textbf{0.00} &  \textbf{0.00} &  \textbf{0.00} \\
    Ablation &  0.13 &  0.11 &  0.07 &  0.10 \\
    Bivariate &  \textbf{0.00} &  \textbf{0.00} &  \textbf{0.00} &  \textbf{0.00} \\
\bottomrule
\label{table:UCI_bank_kendall}
\end{tabular}
\end{table}

For each dataset we show the MKD distances for different length of top-$k$ lists in Tables~\ref{table:UCI_heart_kendall}--\ref{table:UCI_bank_kendall}. In addition, we show the feature importance rankings MCI and Shapley methods produce before and after the duplication in Figures~\ref{fig:heart_dup_experiment}--\ref{fig:bank_dup_experiment}. 

For all datasets, the scores assigned by MCI were practically identical with or without the duplicated features. However the score assigned by the Shapley method to the feature that was duplicated was reduced significantly once duplicated. On the Heart Disease dataset and on the Wine Quality dataset, this was sufficient to change the ranking of the top feature to the $2^{\mbox{nd}}$ or even the $3^{\mbox{rd}}$ position. On the German Credit Default and the Bike Rental datasets the top feature, which had a big margin over the $2^{\mbox{nd}}$ most important feature, maintained it's position but with small margin. On the Online Shopping dataset and the Bank marketing dataset, the original margin of the top feature was so big that even though the score was reduced after duplicating the top feature, it remained in top position with a large margin.

\section{Sensitivity of Different Scores Approximations to Random Seed}
In this section we show results for repeating the quality experiment on the BRCA dataset described in section~$5$ using different random seeds. We show that while the SAGE~\cite{covert2020understanding} method was highly sensitive to the choice of random seed, other methods succeeded to remain stable. Note that in this experiments we used the SAGE method for the global-natural scenario, although SAGE method was originally designed for the global-model scenario.

To run SAGE, we trained a logistic regression model over $2/3$ of the samples and ran SAGE over the remaining $1/3$ with marginal sampling size of 512 as suggested by~\cite{covert2020understanding}. We used the SAGE code available online\footnote{https://github.com/icc2115/sage}.

To measure sensitivity we used the MKD distance (see section~$5$) between ranking prefixes produces by the same method, but with a different random seed. In table~\ref{table:BRCA sensitivity random seeds} we show the mean MDK distance between rankings prefixes produced with different random seeds. In table~\ref{table:BRCA NDCG random seeds} we show the NDCG scores for the rankings prefixes (see section~\ref{Sec:experiments}). As shown in these tables, SAGE was highly sensitive for the choice of random seed while the other methods was highly stable.

\begin{table}[t]
\caption{\bf{Sensitivity of scores to different random seeds}:  \textmd{Here we show the sensitivity of the ranking each score produced to five different random seeds. We measure sensitivity by the average $\pm$ std distance between the ranking produced using each score with different random seeds, for varying ranking prefixes lengths. We measure the distance between each pair of rankings using the MKD distance described in section~\ref{Sec:experiments}. As shown below, the ranking produced by SAGE is highly sensitive to the choice of random seed, while the other scores are insensitive.}}
  \label{table:BRCA sensitivity random seeds}
  \centering
\begin{tabular}{lccccc}
\toprule
Method &              @3 &              @5 &             @10 &             @20 &             @50 \\
\midrule
MCI      &    0.00 $\pm$ 0.00 &    0.00 $\pm$ 0.00 &  0.01 $\pm$ 0.01 &  0.02 $\pm$ 0.01 &   0.02 $\pm$ 0.00 \\
Shapley  &    0.00 $\pm$ 0.00 &    0.00 $\pm$ 0.00 &    0.00 $\pm$ 0.00 &   0.01 $\pm$ 0.00 &    0.00 $\pm$ 0.00 \\
Bivarite &    0.00 $\pm$ 0.00 &    0.00 $\pm$ 0.00 &    0.00 $\pm$ 0.00 &    0.00 $\pm$ 0.00 &    0.00 $\pm$ 0.00 \\
Ablation &    0.00 $\pm$ 0.00 &    0.00 $\pm$ 0.00 &    0.00 $\pm$ 0.00 &    0.00 $\pm$ 0.00 &    0.00 $\pm$ 0.00 \\
SAGE     &  0.09 $\pm$ 0.13 &  0.09 $\pm$ 0.06 &  0.13 $\pm$ 0.06 &  0.12 $\pm$ 0.03 &  0.26 $\pm$ 0.05 \\
\bottomrule
\end{tabular}
\end{table}
\begin{table}[t]
  \caption{\bf{Quality test results for different random seeds}:  \textmd{Below are results of the quality test for the BRCA dataset described in section~\ref{Sec:experiments}, repeated five times with different random seeds. We measure the quality of the ranking each score produced using the NDCG  measure for prefixes of varying sizes, while treating breast cancer related genes as more relevant (higher is better, the perfect score is $1.00$). As shown here, except of the run with random seed $4$, MCI performs substantially better than SAGE.}}
  \label{table:BRCA NDCG random seeds}
  \centering
\begin{tabular}{lcccccc}
\toprule
Method & Seed &        @3  &        @5  &        @10 &        @20 &        @50 \\
\midrule
\multirow{5}{*}{MCI}& 1 &  1.00 &  0.85 &  0.77 &  0.93 &  0.93 \\
%MCI & 1      &  1.00 &  0.85 &  0.77 &  0.93 &  0.93 \\
& 2      &  1.00 &  0.85 &  0.77 &  0.83 &  0.92 \\
%MCI & 2      &  1.00 &  0.85 &  0.77 &  0.83 &  0.92 \\
& 3      &  1.00 &  0.85 &  0.77 &  0.88 &  0.93 \\
%MCI & 3      &  1.00 &  0.85 &  0.77 &  0.88 &  0.93 \\
& 4      &  1.00 &  0.85 &  0.77 &  0.88 &  0.92 \\
%MCI & 4      &  1.00 &  0.85 &  0.77 &  0.88 &  0.92 \\
& 5      &  1.00 &  0.85 &  0.77 &  0.92 &  0.92 \\
%MCI & 5      &  1.00 &  0.85 &  0.77 &  0.92 &  0.92 \\
\hline
\multirow{5}{*}{Shapley} & 1  &  0.77 &  0.70 &  0.73 &  0.88 &  0.88 \\
%Shapley & 1  &  0.77 &  0.70 &  0.73 &  0.88 &  0.88 \\
& 2  &  0.77 &  0.70 &  0.73 &  0.88 &  0.88 \\
& 3  &  0.77 &  0.70 &  0.73 &  0.88 &  0.88 \\
& 4  &  0.77 &  0.70 &  0.73 &  0.88 &  0.88 \\
& 5  &  0.77 &  0.70 &  0.73 &  0.88 &  0.88 \\
\hline
%Shapley & 2  &  0.77 &  0.70 &  0.73 &  0.88 &  0.88 \\
%Shapley & 3  &  0.77 &  0.70 &  0.73 &  0.88 &  0.88 \\
%Shapley & 4  &  0.77 &  0.70 &  0.73 &  0.88 &  0.88 \\
%Shapley & 5  &  0.77 &  0.70 &  0.73 &  0.88 &  0.88 \\
\multirow{5}{*}{Bivarite} & 1 &  1.00 &  0.85 &  0.77 &  0.82 &  0.92 \\
& 2 &  1.00 &  0.85 &  0.77 &  0.82 &  0.92 \\
& 3 &  1.00 &  0.85 &  0.77 &  0.82 &  0.92 \\
& 4 &  1.00 &  0.85 &  0.77 &  0.82 &  0.92 \\
& 5 &  1.00 &  0.85 &  0.77 &  0.82 &  0.92 \\
\hline
%Bivarite & 1 &  1.00 &  0.85 &  0.77 &  0.82 &  0.92 \\
%Bivarite & 2 &  1.00 &  0.85 &  0.77 &  0.82 &  0.92 \\
%Bivarite & 3 &  1.00 &  0.85 &  0.77 &  0.82 &  0.92 \\
%Bivarite & 4 &  1.00 &  0.85 &  0.77 &  0.82 &  0.92 \\
%Bivarite & 5 &  1.00 &  0.85 &  0.77 &  0.82 &  0.92 \\
\multirow{5}{*}{Ablation} & 1 &  0.30 &  0.21 &  0.28 &  0.44 &  0.62 \\
& 2 &  0.30 &  0.21 &  0.28 &  0.44 &  0.62 \\
& 3 &  0.30 &  0.21 &  0.28 &  0.44 &  0.62 \\
& 4 &  0.30 &  0.21 &  0.28 &  0.44 &  0.62 \\
& 5 &  0.30 &  0.21 &  0.28 &  0.44 &  0.62 \\
%Ablation & 1 &  0.30 &  0.21 &  0.28 &  0.44 &  0.62 \\
%Ablation & 2 &  0.30 &  0.21 &  0.28 &  0.44 &  0.62 \\
%Ablation & 3 &  0.30 &  0.21 &  0.28 &  0.44 &  0.62 \\
%Ablation & 4 &  0.30 &  0.21 &  0.28 &  0.44 &  0.62 \\
%Ablation & 5 &  0.30 &  0.21 &  0.28 &  0.44 &  0.62 \\
\hline
\multirow{5}{*}{SAGE}& 1     &  0.70 &  0.79 &  0.58 &  0.70 &  0.83 \\
& 2     &  0.70 &  0.64 &  0.70 &  0.70 &  0.83 \\
& 3     &  0.77 &  0.83 &  0.61 &  0.73 &  0.87 \\
& 4     &  1.00 &  1.00 &  0.87 &  0.92 &  0.97 \\
& 5     &  0.77 &  0.70 &  0.74 &  0.79 &  0.88 \\
%SAGE & 1     &  0.70 &  0.79 &  0.58 &  0.70 &  0.83 \\
%SAGE & 2     &  0.70 &  0.64 &  0.70 &  0.70 &  0.83 \\
%SAGE & 3     &  0.77 &  0.83 &  0.61 &  0.73 &  0.87 \\
%SAGE & 4     &  1.00 &  1.00 &  0.87 &  0.92 &  0.97 \\
%SAGE & 5     &  0.77 &  0.70 &  0.74 &  0.79 &  0.88 \\
\bottomrule
\end{tabular}

\end{table}

%\appendix

%\renewcommand\thesection{\Alph{section}}
%\renewcommand\thesubsection{\thesection.\Roman{subsection}}
%\subimport{./}{proofs}
%\subimport{./}{properties}
%\subimport{./}{computation}
%\subimport{./}{approximations}
%\subimport{./}{UCI experiments}
%\subimport{./}{sensitivity_random_seed}

%\subimport{./}{axioms + why it doesn't fail}
%\subimport{./}{properties}
%\subimport{./}{comparison - real data}
%\subimport{./}{complexity and improvements}

%\section*{References}

%References follow the acknowledgments. Use unnumbered first-level heading for
%the references. Any choice of citation style is acceptable as long as you are
%consistent. It is permissible to reduce the font size to \verb+small+ (9 point)
%when listing the references.
%{\bf Note that the Reference section does not count towards the eight pages of content that are allowed.}
\medskip

%\small
\end{document}